\documentclass[a4paper,11pt,oneside]{article}
\usepackage[latin1]{inputenc}
\usepackage{ngerman}
\usepackage{amsmath}
\usepackage{amssymb}
\usepackage{latexsym}
\usepackage[round]{natbib}
\usepackage{makeidx}
\usepackage{nomencl}
\usepackage{graphicx}
\usepackage[arrow, matrix, curve]{xy}
\usepackage{mathrsfs}
\usepackage{dsfont}   %%% fuer \N, \R etc

\newenvironment{keywords}%
   {\begin{trivlist}\item[]{KEYWORDS:}\ }% oder "Keywords:"
   {\end{trivlist}}
\newenvironment{amsclassification}%
   {\begin{trivlist}\item[]{2000 AMS Classification numbers:}\ }% 
   {\end{trivlist}}

 \newtheorem{theorem}{Theorem}[section]

 \newtheorem{remark}[theorem]{Remark}

 \newtheorem{proposition}[theorem]{Proposition}
 \newtheorem{corollary}[theorem]{Corollary}

\newcommand{\SVM}{S}
\newcommand{\SVMn}{T}
\newcommand{\dddelta}{\varepsilon}
\newcommand{\datenzfv}{\mathds{D}}

\title{Qualitative Robustness of Support Vector Machines
      }
\author{Robert Hable and Andreas Christmann \\ 
        Department of Mathematics \\ University of Bayreuth}
\date{}

\begin{document}

\maketitle

\begin{abstract}
  Support vector machines have attracted much attention
  in theoretical and in applied statistics.
  Main topics of recent interest are consistency, 
  learning rates and robustness. In this article, it is shown
  that support vector machines are qualitatively robust.  
  Since support vector machines can be represented
  by a functional on the set of all probability measures,
  qualitative robustness is proven by showing that this functional is
  continuous with respect to the topology generated by
  weak convergence of probability measures. Combined with the
  existence and uniqueness of support vector machines,
  our results show that support vector machines are the solutions
  of a well-posed mathematical problem in Hadamard's sense. 
\end{abstract}

\begin{amsclassification}
  62G08, 62G35
\end{amsclassification}
\begin{keywords}
  Nonparametric regression, classification,
            machine learning, support vector machines,
            qualitative robustness%
\end{keywords}

\section{A Long Introduction}  \label{sec-introduction}

Two of the most important topics in statistics are 
classification and regression. 
There, it is assumed that
the outcome $y\in\mathcal{Y}$ of a random variable $Y$
(output variable) is influenced by
an observed value $x\in\mathcal{X}$ (input variable). 
On the basis of a finite data set
$\big((x_{1},y_{1}),\dots,(x_{n},y_{n})\big)\in
 (\mathcal{X}\times\mathcal{Y})^{n}$\,, 
the goal is to find an ``optimal'' predictor 
$f:\mathcal{X}\rightarrow\mathcal{Y}$ which
makes a prediction $f(x)$ for an
unobserved $y$\,. In parametric statistics,
a signal plus noise relationship 
$$y\;=\;f_{\theta}(x)+\varepsilon 
$$ 
is often assumed, 
where $f_{\theta}$ is precisely known except for
a finite parameter $\theta\in\mathds{R}^{p}$ and $\varepsilon$ is an
error term (generated from a Normal distribution). 
In this way, the goal of estimating an ``optimal'' predictor
(which can be any function $f:\mathcal{X}\rightarrow\mathcal{Y}$)
reduces to the much simpler task of estimating the 
parameter $\theta\in\mathds{R}^{p}$\,.
Since, in many applications, such strong 
assumptions can hardly be justified, 
nonparametric regression has been developed
which avoids (or at least considerably weakens) such assumptions. 
In statistical machine learning, the method of 
support vector machines has been developed as a method of
nonparametric regression; 
see e.g., \cite{vapnik1998}, \cite{schoelkopf2002},
and \cite{steinwart2008}. There, the estimation of the predictor 
(called \textit{empirical SVM}) is a
function $f$ which solves the minimization problem
\begin{eqnarray}\label{introduction-empirical-svm}
  \min_{f\in H}\,
  \frac{1}{n}\sum_{i=1}^{n} L\big((x_{i},y_{i},f(x_{i})\big)
  \,+\,\lambda\|f\|_{H}^{2}\;,
\end{eqnarray}
where $H$ is a certain function space $H$\,.
The first term in
(\ref{introduction-empirical-svm}) is the empirical mean of
the losses caused by the predictions $f(x_{i})$ and
the second term penalizes the complexity of $f$ in order to
avoid overfitting, $\lambda$ is a positive real number, and
the space $H$ is a reproducing kernel Hilbert space (RKHS)
which consists of
functions $f:\mathcal{X}\rightarrow\mathds{R}$\,.

Since the arise of robust statistics 
(\cite{tukey1960}, \cite{huber1964}),
it is well-known that imperceptible small deviations of the 
real world from 
model assumptions may lead to arbitrarily wrong conclusions.
While many practitioners are aware of the need for robust methods
in classical parametric statistics, it is quite often overseen
that robustness is also a crucial issue in nonparametric statistics.
For example, the sample mean can be seen as a nonparametric 
procedure which is 
non-robust since it is extremely sensitive to outliers: 
Let $X_{1},\dots,X_{n}$ be i.i.d.\ random variables with unknown 
distribution $\mathrm{P}$ and the task is to estimate the 
expectation of $\mathrm{P}$\,.
If the observed data are really generated by the ideal $\mathrm{P}$
(and if expectation and variance of $\mathrm{P}$ exist),
then the sample mean is the optimal estimator. However, 
it frequently happens in the real world that, due to 
outliers or small model violations, 
the observed data are not generated by the ideal
$\mathrm{P}$ but by another distribution $\mathrm{P}^{\prime}$\,. Even if 
$\mathrm{P}^{\prime}$ is close to the ideal $\mathrm{P}$\,, 
the sample mean may lead to disastrous results. 
Detailed descriptions and some examples of such effects
are given, e.g., in \cite{tukey1960}, \cite{huber1964}, and
\citet[\S\,1.1]{huber1981}.

In nonparametric regression, similar effects can occur. There,
it is often assumed that  
$(X_{1},Y_{1}),\dots,(X_{n},Y_{n})$ are 
i.i.d.\ random variables with unknown distribution $\mathrm{P}$\,.
This distribution $\mathrm{P}$ 
determines in which way the output variable $Y_{i}$ 
is influenced by the input variable $X_{i}$. However,
estimating a predictor $f:\mathcal{X}\rightarrow\mathcal{Y}$
can be severely distorted if the
observed data 
$(x_{1},y_{1}),\dots,(x_{n},y_{n})$ are -- just as usual -- 
not generated by
$\mathrm{P}$ but by another distribution $\mathrm{P}^{\prime}$ 
which may be close to the 
ideal $\mathrm{P}$. In order to safeguard from severe distortions,
an estimator $\SVM_{n}$ should fulfill some kind of continuity: 
If the real distribution $\mathrm{P}^{\prime}$ is close to the
ideal distribution $\mathrm{P}$\,, then the distribution of the
estimator $\SVM_{n}$ should hardly be affected (uniformly in 
the sample sizes $n\in\mathds{N}$). 
This kind of robustness is called \textit{qualitative robustness}
and has been formalized in
\citet{hampel1968, hampel1971} for 
estimators taking values in 
$\mathds{R}^{p}$\,. 

In order to study this notion of robust statistics for
support vector machines, we need a generalization
given by \cite{cuevas1988} 
of this formalization 
because, here, the values of the estimator
are \emph{functions} $f:\mathcal{X}\rightarrow\mathcal{Y}$ which are
elements of a (typically infinite dimensional) Hilbert space $H$\,. 
In case of support vector machines, the estimators
$$\SVM_{n}\;:\;\;(\mathcal{X}\times\mathcal{Y})^{n}\;\rightarrow\;H
$$
can be represented by a functional 
$$\SVM\;:\;\;\mathcal{M}_{1}(\mathcal{X}\times\mathcal{Y})
  \;\rightarrow\;H
$$
on the set 
$\mathcal{M}_{1}(\mathcal{X}\times\mathcal{Y})$ of all 
probability measures on $\mathcal{X}\times\mathcal{Y}$\,:
$$\SVM_{n}\big((x_{1},y_{1}),\dots,(x_{n},y_{n})\big)
  \;=\;
  \SVM\Bigg(\frac{1}{n}\sum_{i=1}^{n}\delta_{(x_{i},y_{i})}\Bigg)
$$
for every $(x_{1},y_{1}),\dots,(x_{n},y_{n})\in\mathcal{X}\times\mathcal{Y}$
where $\frac{1}{n}\sum_{i=1}^{n}\delta_{(x_{i},y_{i})}$ is the 
empirical measure and $\delta_{(x_{i},y_{i})}$ denotes the Dirac measure
in $(x_{i},y_{i})$\,.
It is shown by \cite{cuevas1988} that, in such cases,
the qualitative robustness of a sequence of estimators
$(\SVM_{n})_{n\in\mathds{N}}$ follows from the continuity of the
functional $\SVM$ (with respect to the topology of weak convergence of
probability measures). While quantitative
robustness 
of support vector
machines has already been investigated by means of
Hampel's influence functions 
and bounds for the maxbias
in \cite{christmannsteinwart2007})
and by means of 
Bouligand influence functions 
in \cite{christmannvanmessem2008},
results about 
qualitative robustness of support vector machines have not been 
published so far. The goal of this paper is to fill this gap
on research on qualitative robustness of support vector machines.

The structure of the article is as follows: 
In the following Section \ref{subsec-setup},
we recall the basic setup concerning support vector machines, define
the functional $\SVM$ which represents the SVM-estimators
$\SVM_{n}$\,, $n\in\mathds{N}$, and
quote the mathematical definition of qualitative robustness.
In Section \ref{sec-main-results}, we show 
that the functional $\SVM$  of support vector
machines is, in fact, continuous
under very mild assumptions 
(Theorem \ref{theorem-continuity-of-svm-functional}). 
In this way,
it is also proven that, under the same assumptions, support vector 
machines are qualitatively
robust (Theorem \ref{theorem-main-theorem}). 
In addition, it follows that empirical support vector machines
are continuous in the data -- i.e., they
are hardly affected by slight changes in the data
(Corollary \ref{cor-2-theorem-continuity-of-svm-functional}).
Under somewhat different assumptions, this has already been shown in
\citet[Lemma 5.13]{steinwart2008}.
Section \ref{ref-conclusions} contains some concluding remarks.
All proofs are given in the Appendix.

It has to be pointed out that our results 
show
that support vector machines are 
qualitatively robust with a \emph{fixed} regularization 
parameter $\lambda\in(0,\infty)$.
If the fixed regularization parameter $\lambda$ is replaced by
a sequence of parameters $\lambda_{n}\in(0,\infty)$
which decreases to 0 with increasing sample size $n$,
then support vector machines are 
\emph{not} qualitatively robust any more under extremely mild 
conditions. This is demonstrated in Section
\ref{sec-counterexample} in the Appendix.
From our point of view, this is an important result as
all universal consistency proofs we know of for 
support vector machines or for their risks, 
use an appropriate \emph{null sequence} $\lambda_n \in(0,\infty)$, 
$n\in\mathds{N}$.

\section{Support Vector Machines and Qualitative Robustness}
    \label{subsec-setup}

Let $(\Omega,{\cal A},\mathrm{Q})$ be a probability space, 
let $\mathcal{X}$ be a Polish space with Borel-$\sigma$-algebra
$\mathfrak{B}(\mathcal{X})$ and let $\mathcal{Y}$ be a 
closed subset of $\mathds{R}$ with Borel-$\sigma$-algebra
$\mathfrak{B}(\mathcal{Y})$\,. The Borel-$\sigma$-algebra
of $\mathcal{X}\times\mathcal{Y}$ is denoted by
$\mathfrak{B}(\mathcal{X}\times\mathcal{Y})$ and the
set of all probability measures on 
$\big(\mathcal{X}\times\mathcal{Y},
  \mathfrak{B}(\mathcal{X}\times\mathcal{Y})
 \big)
$
is denoted by $\mathcal{M}_{1}(\mathcal{X}\times\mathcal{Y})$\,.
Let
$$X_{1},\dots,X_{n}\;:\;\;(\Omega,{\cal A},\mathrm{Q})
  \;\longrightarrow\;\big(\mathcal{X},\mathfrak{B}(\mathcal{X})\big)
$$
and
$$Y_{1},\dots,Y_{n}\;:\;\;(\Omega,{\cal A},\mathrm{Q})
  \;\longrightarrow\;\big(\mathcal{Y},\mathfrak{B}(\mathcal{Y})\big)
$$
be random variables such that
$\,(X_{1},Y_{1}),\dots,(X_{n},Y_{n})\,$ are independent 
and identically distributed according to some unknown
probability measure
$\mathrm{P}\in\mathcal{M}_{1}(\mathcal{X}\times\mathcal{Y})$\,.

A measurable map
$\,L:\mathcal{X}\times\mathcal{Y}\times\mathds{R}\rightarrow[0,\infty)\,
$
is called \emph{loss function}. It is assumed that
$L(x,y,y)=0$ for every $(x,y)\in\mathcal{X}\times\mathcal{Y}$
-- that is, the loss is zero if the prediction $f(x)$ equals the
observed value $y$\,. 
In addition, we will assume that 
$$L(x,y,\cdot)\;:\;\;\mathds{R}\;\rightarrow\;[0,\infty)\,,\qquad
  t\;\mapsto\;L(x,y,t)
$$
is convex for every $(x,y)\in\mathcal{X}\times\mathcal{Y}$ and
that the following uniform Lipschitz property is fulfilled
for   
a positive real number $|L|_{1}\in(0,\infty)$\,:
\begin{eqnarray}\label{def-uniformly-lipschitz}
  \sup_{(x,y)\in\mathcal{X}\times\mathcal{Y}}
  \big|L(x,y,t)-L(x,y,t^{\prime})\big|\;\;\leq\;\;
  |L|_{1}\cdot|t-t^{\prime}|
  \qquad\; \forall\,t,t^{\prime}\in\mathds{R}\;.
\end{eqnarray}
We restrict our attention to Lipschitz continuous loss functions
because the use of loss functions which are not Lipschitz continuous 
(such as the least squares loss on unbounded domains) 
usually conflicts with 
several notions of robustness; see, e.g., 
% \cite{christmann2004} and 
\citet[\S\,10.4]{steinwart2008}.

The \emph{risk} of a measurable function 
$f:\mathcal{X}\rightarrow\mathds{R}$
is defined by
$$\mathcal{R}_{L,\mathrm{P}}(f)\;=\;
  \int_{\mathcal{X}\times\mathcal{Y}}L\big(x,y,f(x)\big)\,
  \mathrm{P}\big(d(x,y)\big)\;.
$$

Let $k:\mathcal{X}\times\mathcal{X}\rightarrow\mathds{R}$ be a 
bounded and continuous \emph{kernel} with 
\emph{reproducing kernel Hilbert space} (RKHS) $H$. 
See e.g.\ \cite{schoelkopf2002} or \cite{steinwart2008}
for details about these concepts.
Note that 
$H$ is a Polish space
since every Hilbert space is complete and,
according to
\citet[Lemma 4.29]{steinwart2008}, $H$ is separable.
Furthermore,
every $f\in H$ is a bounded and continuous function
$f:\mathcal{X}\rightarrow\mathds{R}$\,;
see \citet[Lemma 4.28]{steinwart2008}. In particular,
every $f\in H$ is measurable and its 
\emph{regularized risk} is defined to be
$$\mathcal{R}_{L,\mathrm{P},\lambda}(f)
  \;=\;\mathcal{R}_{L,\mathrm{P}}(f)\,+\,\lambda\|f\|_{H}^2\;.
$$

An element $f\in H$ is called a \emph{support vector machine} 
and denoted by $f_{L,\mathrm{P},\lambda}$ if it
minimizes the regularized risk in $H$\,. That is,
$$\mathcal{R}_{L,\mathrm{P}}(f_{L,\mathrm{P},\lambda})
      \,+\,\lambda\|f_{L,\mathrm{P},\lambda}\|_{H}^2\;=\;
  \inf_{f\in H}\,\mathcal{R}_{L,\mathrm{P}}(f)\,+\,\lambda\|f\|_{H}^2\;.
$$
We would like to consider a functional
\begin{eqnarray}\label{pre-svm-functional}
  \SVM\;:\;\;\mathrm{P}\;\mapsto\;f_{L,\mathrm{P},\lambda}\;.
\end{eqnarray}
However, support vector machines $f_{L,\mathrm{P},\lambda}$ 
need not exist for every probability measure
$\mathrm{P}\in\mathcal{M}_{1}(\mathcal{X}\times\mathcal{Y})$ and,
therefore, $\SVM$ cannot be defined on 
$\mathcal{M}_{1}(\mathcal{X}\times\mathcal{Y})$
in this way. A sufficient condition for existence of a
support vector machine based on a bounded kernel $k$
is, for example,
$\mathcal{R}_{L,\mathrm{P}}(0)<\infty$;
see \citet[Corollary 5.3]{steinwart2008}. 
In order to enlarge the applicability
of support vector machines, the following extension has been developed
in \cite{christmann2009}.
Following an idea already used by \cite{huber1967} for M-estimates in 
parametric models, a \emph{shifted loss function}
$L^{\ast}:\mathcal{X}\times\mathcal{Y}\times\mathds{R}
  \rightarrow\mathds{R}\,
$
is defined by
$$L^{\ast}(x,y,t)\;=\;L(x,y,t)-L(x,y,0)
  \qquad \forall\,(x,y,t)\in\mathcal{X}\times\mathcal{Y}\times\mathds{R}\;.
$$
Then, similar to the original loss function $L$, define
the $L^{\ast}$\,-\,risk by 
$$\mathcal{R}_{L^{\ast},\mathrm{P}}(f)
     \;=\;\int L^{\ast}\big(x,y,f(x)\big)\,\mathrm{P}\big(d(x,y)\big)
$$ 
and the regularized $L^{\ast}$\,-\,risk by
$$\mathcal{R}_{L^{\ast},\mathrm{P},\lambda}(f)\;=\;
    \mathcal{R}_{L^{\ast},\mathrm{P}}(f)
    \,+\,\lambda\|f\|_{H}^2
$$
for every $f\in H$\,. In complete analogy to 
$f_{L,\mathrm{P},\lambda}$\,, we define the support vector machine
based on the shifted loss function $L^{\ast}$ by
$$f_{L^{\ast},\mathrm{P},\lambda}\;=\;
  \arg\inf_{f\in H}\mathcal{R}_{L^{\ast},\mathrm{P}}(f)+\lambda\|f\|_{H}^2
  \;\;.
$$
The following theorem summarizes some basic results derived by 
\cite{christmann2009}:
\begin{theorem}\label{theorem-L-star-trick-summary}
  For any $\mathrm{P}\in\mathcal{M}_{1}(\mathcal{X}\times\mathcal{Y})$\,,
  there 
  exists a unique $f_{L^{\ast},\mathrm{P},\lambda}\in H$ which minimizes
    $\mathcal{R}_{L^{\ast},\mathrm{P},\lambda}$\,, i.e.
    $$\mathcal{R}_{L^{\ast},\mathrm{P}}(f_{L^{\ast},\mathrm{P},\lambda})
        \,+\,\lambda\|f_{L^{\ast},\mathrm{P},\lambda}\|_{H}^2\;=\;
      \inf_{f\in H}\,\mathcal{R}_{L^{\ast},\mathrm{P}}(f)
          \,+\,\lambda\|f\|_{H}^2\;.
    $$
  If a support vector machine $f_{L,\mathrm{P},\lambda}\in H$ exists 
    (which minimizes $\mathcal{R}_{L,\mathrm{P},\lambda}$ in $H$),
    then 
  $$f_{L^{\ast},\mathrm{P},\lambda}\;=\;f_{L,\mathrm{P},\lambda}\;.
  $$
\end{theorem}
According to this theorem, the map 
$$\SVM\;:\;\;\mathcal{M}_{1}(\mathcal{X}\times\mathcal{Y})
  \;\rightarrow\;H\,,\quad\;
  \mathrm{P}\;\mapsto\;f_{L^{\ast},\mathrm{P},\lambda}
$$
exists, is uniquely defined and
extends the functional in
(\ref{pre-svm-functional}). Therefore, $\SVM$ may be called 
\textit{SVM-functional}.

In order to estimate a measurable map 
$f:\mathcal{X}\rightarrow\mathds{R}$ 
which minimizes the risk
$$\mathcal{R}_{L,\mathrm{P}}(f)\;=\;
  \int_{\mathcal{X}\times\mathcal{Y}}L\big(x,y,f(x)\big)\,
  \mathrm{P}\big(d(x,y)\big)\;,
$$
the \textit{SVM-estimator} is defined by
$$\SVM_{n}\;:\;\;(\mathcal{X}\times\mathcal{Y})^n\;\rightarrow\;H\,,\qquad
  D_{n}\;\mapsto\;
  f_{L,D_{n},\lambda}
$$
where $f_{L,D_{n},\lambda}$ is that function $f\in H$ which minimizes
$$\frac{1}{n}\sum_{i=1}^{n}L\big(x_{i},y_{i},f(x_{i})\big)
  \,+\,\lambda\|f\|_{H}^{2}
$$
in $H$ for 
$D_{n}=((x_{1},x_{2}),\dots,(x_{n},y_{n}))\,\in\,
 (\mathcal{X}\times\mathcal{Y})^{n}
$\,.
Let $\mathds{P}_{D_{n}}$ be the empirical measure corresponding to
the data $D_{n}$ for sample size $n\in\mathds{N}$\,.
Then, the definitions given above yield
\begin{eqnarray}\label{representation-of-empirical-svm}
  f_{L,D_{n},\lambda}\;=\;\SVM_{n}(D_{n})\;=\;
  \SVM(\mathds{P}_{D_{n}})\;=\;f_{L,\mathds{P}_{D_{n}},\lambda}\;\;.
\end{eqnarray}
Note that the support vector machine uniquely exists for every 
empirical measure. In particular, this also implies
$f_{L,D_{n},\lambda}=f_{L^{\ast},\mathds{P}_{D_{n}},\lambda}$\,.
 
\smallskip

\emph{The main goal of the article is to show that,
under very mild conditions, the sequence of
SVM-estimators $(\SVM_{n})_{n\in\mathds{N}}$ is
qualitatively robust.} According to \citet[Definition 1]{cuevas1988},
the sequence $(\SVM_{n})_{n\in\mathds{N}}$ is called 
\emph{qualitatively robust} if
the functions
$$\mathcal{M}_{1}(\mathcal{X}\times\mathcal{Y})\;\rightarrow\;
  \mathcal{M}_{1}(H)\,,\qquad
  \mathrm{P}\;\mapsto\;\SVM_{n}(\mathrm{P}^n)\;,\qquad n\in\mathds{N}\;,
$$
are uniformly continuous with respect to the weak
topologies 
on $\mathcal{M}_{1}(\mathcal{X}\times\mathcal{Y})$
and $\mathcal{M}_{1}(H)$\,.
Here, $\mathcal{M}_{1}(H)$ denotes the set of all probability measures
on $(H,\mathfrak{B}(H))$\,, $\mathfrak{B}(H)$ is the
Borel-$\sigma$-algebra on $H$, and $\SVM_{n}(\mathrm{P}^n)$ denotes
the image measure of $\mathrm{P}^n$ with respect to $\SVM_{n}$\,.
Hence, $\SVM_{n}(\mathrm{P}^n)$ is the measure on $(H,\mathfrak{B}(H))$
which is defined by
$$\big(\SVM_{n}(\mathrm{P}^n)\big)(F)\;=\;
  \mathrm{P}^n\Big(\big\{D_{n}
                \in(\mathcal{X}\times\mathcal{Y})^n\;
          \big|\;\;\SVM_{n}(D_{n})
               \;\in\;F
          \big\}
     \Big)
$$ 
for every Borel-measurable subset $F\subset H$\,.
Of course, this definition only makes sense if the
SVM-estimators are measurable with respect to the 
Borel-$\sigma$-algebras. This measurability is
assured by Corollary 
\ref{cor-2-theorem-continuity-of-svm-functional} below.

Since the weak topologies on 
$\mathcal{M}_{1}(\mathcal{X}\times\mathcal{Y})$
and $\mathcal{M}_{1}(H)$ 
are metrizable by the Prokhorov metric 
$d_{\textrm{Pro}}$ (see Subsection \ref{sec-weak-convergence-bochner}), 
the sequence of
SVM-estimators $(\SVM_{n})_{n\in\mathds{N}}$ is
qualitatively robust if and only if for every 
$\mathrm{P}\in\mathcal{M}_{1}(\mathcal{X}\times\mathcal{Y})$ and every 
$\rho>0$ there is an $\varepsilon>0$ such that
$$ % \mathrm{Q}\in\mathcal{M}_{1}(\mathcal{X}\times\mathcal{Y}),\;\,
  d_{\textrm{Pro}}(\mathrm{Q},\mathrm{P})\,<\,\varepsilon
  \quad\Rightarrow\quad
  d_{\textrm{Pro}}\big(\SVM_{n}(\mathrm{Q}^n),\SVM_{n}(\mathrm{P}^n)\big)
  \,<\,\rho
    \;\;\forall\,n\in\mathds{N}\;.
$$

\begin{figure}
  \begin{center}
    \includegraphics[width=0.95\textwidth,angle=0]{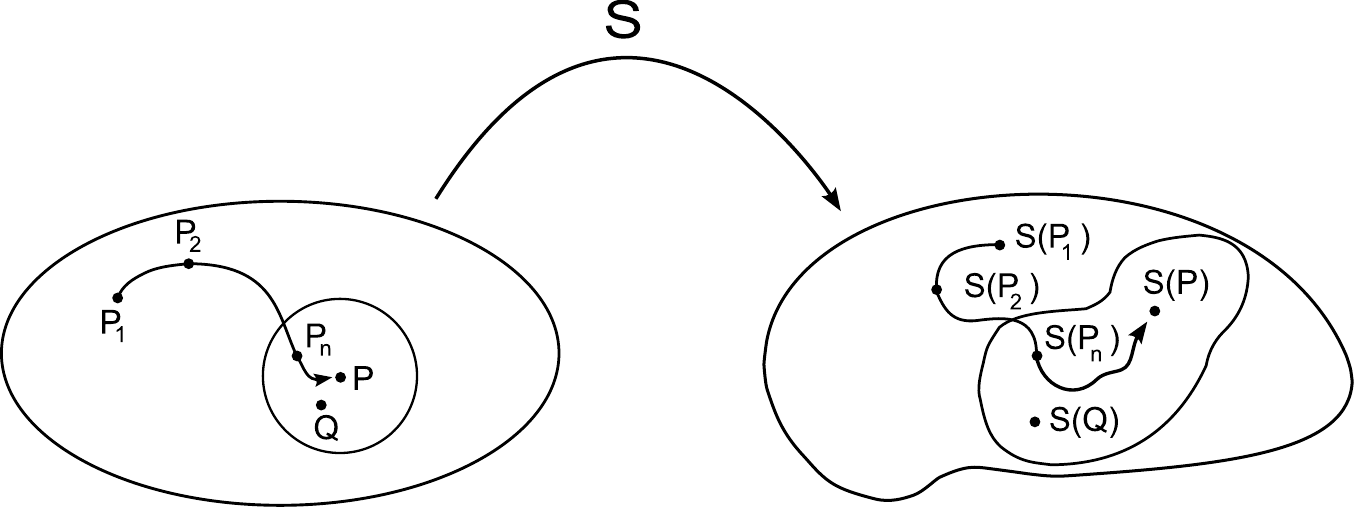}
  \end{center}
  \caption{Sketch: reasoning of robustness of $\SVM(\mathrm{P})$. 
           Left: $\mathrm{P}$, a neighborhood of $\mathrm{P}$, 
                 and $\mathcal{M}_{1}(\mathcal{X}\times\mathcal{Y})$.
           Right: $\SVM(\mathrm{P})$, a neighborhood of 
                $\SVM(\mathrm{P})$, and
                the space of all probability
                measures of $\SVM(\mathrm{P})$ for 
                $\mathrm{P}\in\mathcal{M}_{1}(\mathcal{X}\times\mathcal{Y})$.
  \label{robsec:1:scheme}}
\end{figure}

Roughly speaking, qualitative robustness means that the SVM-estimator
tolerates two kinds of errors in the data: small errors
in many observations $(x_{i},y_{i})$ and
large errors in a small fraction of the data set. 
These two kinds of errors only have slight effects on 
the distribution and, therefore, on the
performance of the SVM-estimator (uniformly in the sample size).
Figure \ref{robsec:1:scheme} gives a graphical illustration
of qualitative robustness.

\section{Main Results}\label{sec-main-results}

The following theorem is our main result and shows
that support vector machines are qualitatively robust
under mild conditions. 
\begin{theorem}\label{theorem-main-theorem}
  Let $\mathcal{X}$ be a Polish space and let $\mathcal{Y}$ be a 
  closed subset of $\mathds{R}$\,. Let the loss function
  be a continuous function
  $\,L:\mathcal{X}\times\mathcal{Y}\times\mathds{R}\rightarrow[0,\infty)\,
  $
  such that $L(x,y,y)=0$ for every $(x,y)\in\mathcal{X}\times\mathcal{Y}$
  and
  $$L(x,y,\cdot)\;:\;\;\mathds{R}\;\rightarrow\;[0,\infty)\,,\qquad
    t\;\mapsto\;L(x,y,t)
  $$
  is convex for every $(x,y)\in\mathcal{X}\times\mathcal{Y}$\,.
  Assume that
  the uniform Lipschitz property 
  \begin{eqnarray*}
    \sup_{(x,y)\in\mathcal{X}\times\mathcal{Y}}
    \big|L(x,y,t)-L(x,y,t^{\prime})\big|\;\;\leq\;\;
    |L|_{1}\cdot|t-t^{\prime}|
    \qquad\; \forall\,t,t^{\prime}\in\mathds{R}
  \end{eqnarray*}
  is fulfilled
  for a real number $|L|_{1}\in(0,\infty)$\,. Furthermore, 
  let $k:\mathcal{X}\times\mathcal{X}\rightarrow\mathds{R}$ be a 
  bounded and continuous kernel with RKHS $H$\,.
  
  Then, the sequence of
  SVM-estimators $(\SVM_{n})_{n\in\mathds{N}}$ is
  qualitatively robust.
\end{theorem}
Of course, this theorem applies to classification
(e.g.\ $\mathcal{Y}=\{-1,1\}$) and 
regression (e.g.\ $\mathcal{Y}=\mathds{R}$ or $\mathcal{Y}=[0,\infty)$).
In particular, note that every function 
$g:\mathcal{Y}\rightarrow\mathds{R}$ is continuous if $\mathcal{Y}$
is a discrete set -- e.g.\ $\mathcal{Y}=\{-1,1\}$\,.
In this case, assuming $L$ to be continuous reduces
to the assumption that
$$\mathcal{X}\times\mathds{R}\;\rightarrow\;[0,\infty)\,,\qquad
    (x,t)\;\mapsto\;L(x,y,t)
$$
is continuous for every $y\in\mathcal{Y}$\,.
Many of the most common loss functions are permitted in the theorem,
e.g.\ the hinge loss and logistic loss for classification, 
$\varepsilon$-insensitive loss and Huber's loss for regression, 
and the pinball loss for quantile regression. 
The least squares loss is ruled out in Theorem
\ref{theorem-main-theorem}  
-- which is not surprising as it is the prominent
standard example of a loss function which typically conflicts with
robustness if $\mathcal{X}$ and $\mathcal{Y}$ are unbounded;
see, e.g., \cite{christmannsteinwart2007} and 
\cite{christmannvanmessem2008}.
Assuming continuity of the kernel $k$ does not seem to be 
very restrictive as all of the most common kernels are continuous.
Assuming $k$ to be bounded is quite natural in order to 
ensure good robustness properties. While the
Gaussian RBF kernel is always bounded, polynomial kernels
(except for the constant kernel) and the exponential kernel are 
bounded if and only if $\mathcal{X}$ is bounded.
 
\smallskip

In our definition of the sequence $(\SVM_{n})_{n\in\mathds{N}}$
of SVM-estimators, the regularization parameter $\lambda$ 
is a fixed real number which does not
change with $n$\,. Instead, it is also common to consider
sequences of estimators
$$\SVMn_{n}\;:\;\;(\mathcal{X}\times\mathcal{Y})^{n}\;\rightarrow\;H\,,
    \qquad D_{n}\;\mapsto\;f_{L,D_{n},\lambda_{n}}\,,
    \qquad n\in\mathds{N}\,,
$$
where the fixed parameter $\lambda$ is replaced by
a sequence $(\lambda_{n})_{n\in\mathds{N}}\subset(0,\infty)$
with $\lim_{n\rightarrow\infty}\lambda_{n}=0$\,.
However, Theorem \ref{theorem-main-theorem} \emph{cannot} 
be generalized
to $(\SVMn_{n})_{n\in\mathds{N}}$\,.  
Proposition \ref{prop-counterexample} (in the Appendix)
shows 
under extremely mild conditions
that $(\SVMn_{n})_{n\in\mathds{N}}$ is \emph{not} qualitatively robust.
This is of interest because appropriately chosen null sequences 
$(\lambda_{n})_{n\in\mathds{N}}\subset(0,\infty)$ 
are used to prove universal consistency of
the risk
$\,\mathcal{R}_{L^{\ast},\mathrm{P}}(f_{L^{\ast},D_{n},\lambda_{n}})
 \xrightarrow[]{\,\mathrm{P}}
 \inf_{f\in \mathcal{F}}\mathcal{R}_{L^{\ast},\mathrm{P}}(f)
$\, and
$\,f_{L^{\ast},D_{n},\lambda_{n}}
 \xrightarrow[]{\,\mathrm{P}}
 \textup{arg}\inf_{f\in \mathcal{F}}\mathcal{R}_{L^{\ast},\mathrm{P}}(f)
$\, for $n\rightarrow\infty$
where $\mathcal{F}$ denotes the set of all \emph{measurable}
functions $f:\mathcal{X}\rightarrow\mathds{R}$.
This was first shown by 
\cite{steinwart2002},
\cite{zhang2004b},
and
\cite{steinwart2005}.
We also refer to
\cite{bousquet2002},
\cite{bartlett2006},
\cite{christmann2009}, and
\cite{steinwart2009}.
 
\smallskip

The proof of Theorem \ref{theorem-main-theorem} 
is based on the following result
which is interesting on its own.
\begin{theorem}\label{theorem-continuity-of-svm-functional}
  Under the assumptions of Theorem \ref{theorem-main-theorem},
  the SVM-func\-tional
  $$\SVM\;:\;\;\mathcal{M}_{1}(\mathcal{X}\times\mathcal{Y})
    \;\rightarrow\;H\,,\quad\;
    \mathrm{P}\;\mapsto\;f_{L^{\ast},\mathrm{P},\lambda}
  $$
  is continuous with respect to the weak topology on
  $\mathcal{M}_{1}(\mathcal{X}\times\mathcal{Y})$ and the norm 
  topology on $H$\,.
\end{theorem} 
As a generalization of earlier results by, e.g.,
\citet{Zhang01a}, \citet{DeRoCaPiVe04a}, and \citet{Ste03c},
\citet[Theorem 7]{christmann2009} derived a
representer theorem which showed that, for \emph{every}
$\mathrm{P}_{0}\in\mathcal{M}_{1}(\mathcal{X}\times\mathcal{Y})$,
there is a bounded map 
$h:\mathcal{X}\times\mathcal{Y}\rightarrow\mathds{R}$ such that
$f_{L^{\ast},\mathrm{P}_{0},\lambda}=
  -\frac{1}{2\lambda}\int h\Phi \,d\mathrm{P}_{0}
$\,
and
\begin{eqnarray}\label{remark-theorem-continuity-of-svm-functional-1}
  \big\|f_{L^{\ast},\mathrm{P},\lambda}-f_{L^{\ast},\mathrm{P}_{0},\lambda}
  \big\|_{H}
  \;\leq\;\lambda^{-1}
     \left\| \int h\Phi \,d\mathrm{P} - \int h\Phi \,d\mathrm{P}_{0} 
     \right\|
 % XXX \quad\forall\,P\in\mathcal{M}_{1}(\mathcal{X}\times\mathcal{Y})
\end{eqnarray}
for every $\mathrm{P}\in\mathcal{M}_{1}(\mathcal{X}\times\mathcal{Y})$\,.
The integrals in (\ref{remark-theorem-continuity-of-svm-functional-1})
are Bochner integrals of
the vector-valued function 
$h\Phi:\mathcal{X}\times\mathcal{Y}\rightarrow H$\,, 
$(x,y)\mapsto h(x,y)\Phi(x)$
where $\Phi$ is the canonical feature map of $k$\,, i.e.\
$\Phi(x)=k(\cdot,x)$ for all $x\in\mathcal{X}$\,.
This offers an elegant possibility of proving 
Theorem \ref{theorem-continuity-of-svm-functional}
if we would accept some additional assumptions:
The statement of 
Theorem \ref{theorem-continuity-of-svm-functional} is
true if $\int h\Phi\,d\mathrm{P}_{n}$ converges to
$\int h\Phi\,d\mathrm{P}_{0}$ for every weakly convergent sequence 
$\mathrm{P}_{n}\rightarrow \mathrm{P}_{0}$\,.
In the following, we show that the
integrals indeed converge -- under the additional
assumptions that the 
derivative $\frac{\partial L}{\partial t}(x,y,t)$
exists
and is continuous for every 
$(x,y,t)\in\mathcal{X}\times\mathcal{Y}\times\mathds{R}$\,. 
These assumptions are fulfilled e.g.\ for the logistic 
loss function and Huber's loss function.
In this case, 
it follows from \citet[Theorem 7]{christmann2009}
that $h$ is continuous. Since $\Phi$ is continuous and bounded
(see e.g.\ \citet[p.\ 124 and Lemma 4.29]{steinwart2008},
the integrand $h\Phi:\mathcal{X}\times\mathcal{Y}\rightarrow H$   
is continuous and bounded. Then, it follows from
\citet[p.\ III.40]{bourbaki2004integration} that
$\int h\Phi\,d\mathrm{P}_{n}$ converges to
$\int h\Phi\,d\mathrm{P}_{0}$ for every weakly convergent sequence 
$\mathrm{P}_{n}\rightarrow \mathrm{P}_{0}\,$ --- 
just as in case of real-valued integrands;
see Subsection \ref{sec-weak-convergence-bochner} in the Appendix.

Unfortunately, this short proof only works under the
additional assumption of a continuous 
partial derivative $\frac{\partial L}{\partial t}$ and this assumption 
rules out many loss functions used in practice, such as
hinge, absolute distance and $\varepsilon$-insensitive 
for regression and pinball for quantile regression.
Therefore, our proof of Theorem \ref{theorem-continuity-of-svm-functional}
(without this additional assumption) does not
use the representer theorem and Bochner integrals;
it is mainly based on the theory of Hilbert spaces
and weak convergence of measures. In the following, we give 
some corollaries
of Theorem \ref{theorem-continuity-of-svm-functional}.

\smallskip

Let $\mathcal{C}_{b}(\mathcal{X})$ be the Banach space of
all bounded, continuous functions $f:\mathcal{X}\rightarrow\mathds{R}$
with norm
$$\|f\|_{\infty}\;=\;\sup_{x\in\mathcal{X}}|f(x)|\;.
$$
Since $k$ is continuous and bounded, we immediately get
from Theorem \ref{theorem-continuity-of-svm-functional}
and \citet[Lemma 4.28]{steinwart2008}:
\begin{corollary}\label{cor-1-theorem-continuity-of-svm-functional}
  Under the assumptions of Theorem \ref{theorem-main-theorem},
  the SVM-functional
  $$\mathcal{M}_{1}(\mathcal{X}\times\mathcal{Y})
    \;\rightarrow\;\mathcal{C}_{b}(\mathcal{X})\,,\quad\;
    \mathrm{P}\;\mapsto\;f_{L^{\ast},\mathrm{P},\lambda}
  $$
  is continuous with respect to the weak topology on
  $\mathcal{M}_{1}(\mathcal{X}\times\mathcal{Y})$ and the norm 
  topology on $\mathcal{C}_{b}(\mathcal{X})$\,.

% XXX  
\textup{That is, 
$\,\sup_{x\in\mathcal{X}}
  \big|f_{L,\mathrm{P}^{\prime},\lambda}(x)-f_{L,\mathrm{P},\lambda}(x)
  \big|\,
$
is small
if $\mathrm{P}^{\prime}$ is close to $\mathrm{P}$\,.}
% XXX
\end{corollary} 

The next corollary is similar to \citet[Lemma 5.13]{steinwart2008}
but only assumes continuity instead of differentiability of 
$t\mapsto L(x,y,t)$. In combination with existence and uniqueness
of support vector machines 
(see Theorem \ref{theorem-L-star-trick-summary}),
this result shows that a support vector machine 
is the solution of a well-posed 
mathematical problem in the sense of \cite{hadamard1902}.
\begin{corollary}\label{cor-2-theorem-continuity-of-svm-functional}
  Under the assumptions of Theorem \ref{theorem-main-theorem},
  the SVM-estimator 
  $$\SVM_{n}\;:\;\;(\mathcal{X}\times\mathcal{Y})^n\;\rightarrow\;H\,,
    \qquad
    D_{n}\;\mapsto\;
    f_{L,D_{n},\lambda}
  $$
  is continuous. 
\end{corollary}
In particular, it follows from Corollary 
    \ref{cor-2-theorem-continuity-of-svm-functional} that
    the SVM-estimator $\SVM_{n}$ is measurable.
\begin{remark}
    Let $d_{n}$ be a metric which generates the
    topology on $(\mathcal{X}\times\mathcal{Y})^{n}$\,,
    e.g.\ the Euclidean metric on $\mathds{R}^{n(k+1)}$
    if $\mathcal{X}\subset\mathds{R}^{k}$\,.
    Then  
    Corollary \ref{cor-2-theorem-continuity-of-svm-functional} 
    and \citet[Lemma 4.28]{steinwart2008} imply the following
    continuity property of the SVM-estimator:
    For every $\varepsilon>0$
    and every data set 
    $D_{n}\in(\mathcal{X}\times\mathcal{Y})^{n}$, 
    there is a $\delta>0$ such that
    $$\sup_{x\in\mathcal{X}}
      \big|f_{L,D_{n}^{\prime},\lambda}(x)
         -f_{L,D_{n},\lambda}(x)
      \big|\;<\;\varepsilon
    $$
    if $D_{n}^{\prime}\in(\mathcal{X}\times\mathcal{Y})^{n}$ 
    is any other data set with $n$ observations and
    $\,d_{n}(D_{n}^{\prime},D_{n})<\delta$.
\end{remark}

We finish this section with a corollary about strong consistency
of support vector machines which arises as a
by-product of Theorem \ref{theorem-continuity-of-svm-functional}.
Often, asymptotic results of support vector machines
show the convergence in probability of the risk
$\mathcal{R}_{L^{\ast},\mathrm{P}}(f_{L^{\ast},\datenzfv_{n},\lambda_{n}})$
to the \emph{Bayes risk} 
$\inf_{f\in \mathcal{F}}\mathcal{R}_{L^{\ast},\mathrm{P}}(f)$
and of $f_{L^{\ast},\datenzfv_{n},\lambda_{n}}$ to
$\textup{arg}\inf_{f\in \mathcal{F}}\mathcal{R}_{L^{\ast},\mathrm{P}}(f)$\,,
where $\mathcal{F}$ is the set of all measurable
functions $f:\mathcal{X}\rightarrow\mathds{R}$
and $(\lambda_{n})_{n\in\mathds{N}}$ is a suitable
null sequence. In contrast to that,
the following corollary 
provides for \emph{fixed} $\lambda\in(0,\infty)$
almost sure convergence of 
$\mathcal{R}_{L^{\ast},\mathrm{P}}(f_{L^{\ast},\datenzfv_{n},\lambda})$
to 
$\mathcal{R}_{L^{\ast},\mathrm{P}}(f_{L^{\ast},\mathrm{P},\lambda})$
and of $f_{L^{\ast},\datenzfv_{n},\lambda}$ to
$f_{L^{\ast},\mathrm{P},\lambda}$\,.
This is an interesting fact, although 
the limit 
$\mathcal{R}_{L^{\ast},\mathrm{P}}(f_{L^{\ast},\mathrm{P},\lambda})$
will in general differ from the Bayes risk.

Recall from Section \ref{subsec-setup} that the data points
$(x_{i},y_{i})$ from the data set 
$D_{n}=\big((x_{1},x_{2}),\dots(x_{n},y_{n})\big)$
are realizations of i.i.d.\ random variables
$$(X_{i},Y_{i})\;:\;\;(\Omega,{\cal A},\mathrm{Q})
  \;\longrightarrow\;
  \big(\mathcal{X}\times\mathcal{Y},
       \mathfrak{B}(\mathcal{X}\times\mathcal{Y})
  \big)\;,\qquad
    n\in\mathds{N}\,,
$$
such that
$$(X_{i},Y_{i})\;\sim\;\mathrm{P}\qquad\quad\forall\,n\in\mathds{N}\;.
$$
\begin{corollary}\label{cor-consistency}
  Define the random vectors
  $$\datenzfv_{n}\;:=\;\big((X_{1},Y_{1}),\dots,(X_{n},Y_{n})\big)
  $$
  and the corresponding $H$-valued random functions
  $$f_{L^{\ast},\datenzfv_{n},\lambda}\;=\;
    \textup{arg}\inf_{f\in H}\,
    \frac{1}{n}\sum_{i=1}^{n}L^{\ast}\big(X_{i},Y_{i},f(X_{i})\big)
      \,+\,\lambda\|f\|_{H}^{2}\;,\qquad
    n\in\mathds{N}\,.
  $$  
  From the assumptions of Theorem \ref{theorem-main-theorem},
  it follows that 
  
  \smallskip
  
  (a) ${\displaystyle \;
       \lim_{n\rightarrow\infty}
          \|f_{L^{\ast},\datenzfv_{n},\lambda}
              -f_{L^{\ast},\mathrm{P},\lambda}\|_{H}
       \;=\;0\qquad\text{almost sure}}
     $

  \smallskip
  
  (b) ${\displaystyle \;
       \lim_{n\rightarrow\infty}\,\sup_{x\in\mathcal{X}}
          |f_{L^{\ast},\datenzfv_{n},\lambda}(x)
              -f_{L^{\ast},\mathrm{P},\lambda}(x)|
       \;=\;0\qquad\text{almost sure}}
     $

  \smallskip
  
  (c) ${\displaystyle \;
       \lim_{n\rightarrow\infty}\,
          \mathcal{R}_{L^{\ast},\mathrm{P},\lambda}
                 (f_{L^{\ast},\datenzfv_{n},\lambda})
          \;=\;\mathcal{R}_{L^{\ast},\mathrm{P},\lambda}
                    (f_{L^{\ast},\mathrm{P},\lambda})
       \qquad\text{almost sure}}
     $

  \smallskip
  
  (d) ${\displaystyle \;
       \lim_{n\rightarrow\infty}\,
          \mathcal{R}_{L^{\ast},\mathrm{P}}(f_{L^{\ast},\datenzfv_{n},\lambda})
          \;=\;\mathcal{R}_{L^{\ast},\mathrm{P}}
                  (f_{L^{\ast},\mathrm{P},\lambda})
       \qquad\text{almost sure.}}
     $
  
  \smallskip
  
  If the support vector machine $f_{L,\mathrm{P},\lambda}$ exists, 
   then assertions
  (a)--(d) are also valid for $L$ instead of $L^{\ast}$\,.
\end{corollary}

\section{Conclusions} \label{ref-conclusions}

It is well-known that outliers in data sets or other moderate 
model violations can pose a serious problem to a statistical analysis. 
On the one hand, practitioners can hardly guarantee that their data 
sets do not contain any outliers, while, on the other hand, many 
statistical
methods are very sensitive even to small violations of the assumed 
statistical model.
Since support vector machines play an important role in statistical
machine learning, investigating their performance in the presence
of moderate model violations is a crucial topic
-- the more so as support vector machines are
frequently applied to large and complex high-dimensional data sets.  

In this article, we showed that support vector machines are 
qualitatively robust with a fixed regularization 
parameter $\lambda\in(0,\infty)$, i.e., the performance
of support vector machines is hardly affected by the following
two kinds of errors: large errors in a small fraction of the
data set and small errors in the whole data set.
This not only means that these errors do not lead to large
errors in the support vector machines but also that even the
finite sample distribution of support vector machines is 
hardly affected.

In contrast to that, we also showed that support vector machines are 
\emph{not} qualitatively robust any more under extremely mild 
conditions, if the fixed regularization parameter $\lambda$ is replaced by
a sequence of parameters $\lambda_{n}\in(0,\infty)$
which decreases to 0 with increasing sample size $n$. 
From our point of view, this is an important result as
all universal consistency proofs we know of for 
support vector machines or for their risks, 
use an appropriate \emph{null sequence} $\lambda_n \in(0,\infty)$, 
$n\in\mathds{N}$.

\section{Appendix} \label{sec-appendix}

In Subsection \ref{sec-weak-convergence-bochner}, we 
briefly recall some facts about weak convergence
of probability measures. In addition, we show that
weak convergence of probability measures on a Polish space
implies convergence of the corresponding Bochner integrals
of bounded, continuous functions.
Subsection \ref{sec-counterexample} demonstrates 
under extremely mild conditions that
the sequence of SVM-estimators  
cannot be qualitatively robust if the fixed 
regularization parameter $\lambda$ is replaced by
a sequence $(\lambda_{n})_{n\in\mathds{N}}\subset(0,\infty)$
with $\lim_{n\rightarrow\infty}\lambda_{n}=0$\,.
Subsection \ref{sec-proofs}
contains all proofs.

\subsection{Weak Convergence of Probability Measures and 
   Bochner Integrals}  \label{sec-weak-convergence-bochner}

Let $\mathcal{Z}$ be a Polish space with Borel-$\sigma$-algebra
$\mathfrak{B}(\mathcal{Z})$, let $d$ be a metric on
$\mathcal{Z}$ which generates the topology on $\mathcal{Z}$
and let $\mathcal{M}_{1}(\mathcal{Z})$ be the set of all probability
measures on $(\mathcal{Z},\mathfrak{B}(\mathcal{Z}))$\,.

A sequence $(\mathrm{P}_{n})_{n\in\mathds{N}}$
of probability measures on $\mathcal{Z}$ converges
to a probability measure $\mathrm{P}_{0}$ in the
weak topology on $\mathcal{M}_{1}(\mathcal{Z})$ if
$$\lim_{n\rightarrow\infty}\int g \,d\mathrm{P}_{n}
  \;=\;\int g \,d\mathrm{P}_{0}
  \qquad\forall\,g\in\mathcal{C}_{b}(\mathcal{Z})
$$
where $\mathcal{C}_{b}(\mathcal{Z})$ denotes the set of all
bounded, continuous functions $g:\mathcal{Z}\rightarrow\mathds{R}$\,,
see \citet[\S\,1]{billingsley1968}.

The weak topology on $\mathcal{M}_{1}(\mathcal{Z})$ is metrizable 
by the Prokhorov metric 
$d_{\textrm{Pro}}$\,; see e.g.\ \citet[\S\,2.2]{huber1981}.
The Prokhorov metric $d_{\textrm{Pro}}$ on $\mathcal{M}_{1}(\mathcal{Z})$
is defined by
$$d_{\textrm{Pro}}(\mathrm{P}_{1},\mathrm{P}_{2})\;=\;
  \inf\big\{\varepsilon\in(0,\infty)\;
      \big|\;\;\mathrm{P}_{1}(B)\,<\,
               \mathrm{P}_{2}(B^{\varepsilon})+\varepsilon
               \;\;\forall\,B\in\mathfrak{B}(\mathcal{Z})
      \big\}
$$
where 
$B^{\varepsilon}
 =\{z\in\mathcal{Z}\,|\,\inf_{z^{\prime}\in\mathcal{Z}}
                         d(z,z^{\prime})<\varepsilon
  \}
$\,.

Let $g:\mathcal{Z}\rightarrow\mathds{R}$ be a continuous and 
bounded function.
By definition, we have
$\lim_{n\rightarrow\infty}\int g \,d\mathrm{P}_{n}
  =\int g \,d\mathrm{P}_{0}
$
for every sequence 
$(\mathrm{P}_{n})_{n\in\mathds{N}}\subset\mathcal{M}_{1}(\mathcal{Z})$
which converges weakly in $\mathcal{M}_{1}(\mathcal{Z})$
to some $\mathrm{P}_{0}$\,.
The following theorem states that this is still valid 
for Bochner integrals if
$g$ is replaced by a vector-valued continuous and bounded function
$\Psi:\mathcal{Z}\rightarrow H$\,, where $H$ is a 
separable Banach space. 
This follows from 
a corresponding statement in 
\citet[p.\ III.40]{bourbaki2004integration}
for locally compact spaces $\mathcal{Z}$.
Boundedness of $\Psi$ means that 
$\sup_{z\in\mathcal{Z}}\|\Psi(z)\|_{H}<\infty$\,.

\begin{theorem}\label{theorem-continuity-bochner}
  Let $\mathcal{Z}$ be a Polish space with 
  Borel-$\sigma$-algebra $\mathfrak{B}(\mathcal{Z})$ and let
  $H$ be a separable Banach space. If $\Psi:\mathcal{Z}\rightarrow H$
  is a continuous and bounded function, then
  $$\int \Psi\,d\mathrm{P}_{n}\;\;\longrightarrow\;\;
    \int \Psi\,d\mathrm{P}_{0}
    \qquad(n\rightarrow\infty)
  $$
  for every sequence 
  $(\mathrm{P}_{n})_{n\in\mathds{N}}\subset\mathcal{M}_{1}(\mathcal{Z})$
  which converges weakly in $\mathcal{M}_{1}(\mathcal{Z})$
  to some $\mathrm{P}_{0}$\,.
\end{theorem}

\subsection{A Counterexample}\label{sec-counterexample}

Theorem \ref{theorem-main-theorem} shows that,
for a \emph{fixed} regularization parameter $\lambda\in(0,\infty)$\,, 
the sequence of SVM-estimators  
$$\SVM_{n}\;:\;\;(\mathcal{X}\times\mathcal{Y})^{n}\;\rightarrow\;H\,,
    \qquad D_{n}\;\mapsto\;f_{L,D_{n},\lambda}\,,
    \qquad n\in\mathds{N}\,,
$$
is qualitatively robust. The following proposition shows that,
under extremely mild conditions,
the sequence of estimators
$$\SVMn_{n}\;:\;\;(\mathcal{X}\times\mathcal{Y})^{n}\;\rightarrow\;H\,,
    \qquad D_{n}\;\mapsto\;f_{L,D_{n},\lambda_{n}}\,,
    \qquad n\in\mathds{N}\,,
$$
\emph{cannot} be qualitatively robust if the fixed parameter 
$\lambda$ is replaced by
a sequence $(\lambda_{n})_{n\in\mathds{N}}\subset(0,\infty)$
with $\lim_{n\rightarrow\infty}\lambda_{n}=0$\,.
This shows that the asymptotic results on 
\emph{universal consistency} of support vector machines --
which consider appropriate null sequences
$(\lambda_{n})_{n\in\mathds{N}}\subset(0,\infty)$ --
are in conflict with qualitative robustness of
support vector machines using $\lambda_{n}$\,.
(Asymptotic results on 
\emph{universal consistency} of support vector machines
can be found, e.g., in the references listed before
Theorem \ref{theorem-continuity-of-svm-functional}.)

For simplicity,
the following proposition focuses on regression because
it is assumed
that $\{0,1\}\subset\mathcal{Y}$\,.
A similar proposition (with a similar proof) 
can also be given
in case of binary classification where $\mathcal{Y}=\{-1,1\}$\,.

\begin{proposition}\label{prop-counterexample}
  Let $\mathcal{X}$ be a Polish space and let 
  $\mathcal{Y}$ be a closed subset of $\mathds{R}$ such that
  $\{0,1\}\subset\mathcal{Y}$\,.
  Let $k$ be a bounded kernel with RKHS $H$\,.
  Let $L$ be a convex loss function such that 
  $L(x,y,y)=0$ for every $(x,y)\in\mathcal{X}\times\mathcal{Y}$\,.
  In addition, assume that there are $x_{0},x_{1}\in\mathcal{X}$
  such that
  \begin{eqnarray}
   && \exists\,\tilde{f}\in H\,:\quad \tilde{f}(x_{0})=0\,,\quad
      \tilde{f}(x_{1})\not= 0 \label{prop-counterexample-a}\\
   && L(x_{1},1,0)\;>\;0\;. \label{prop-counterexample-b}
  \end{eqnarray}
  Let $(\lambda_{n})_{n\in\mathds{N}}\subset(0,\infty)$ 
  be any sequence
  such that $\lim_{n\rightarrow\infty}\lambda_{n}=0$\,.
  Then, the sequence of estimators
  $$\SVMn_{n}\;:\;\;(\mathcal{X}\times\mathcal{Y})^{n}\;\rightarrow\;H\,,
    \qquad D_{n}\;\mapsto\;f_{L,D_{n},\lambda_{n}}\,,
    \qquad n\in\mathds{N}\,,
  $$
  is not qualitatively robust. 
\end{proposition}

\subsection{Proofs} \label{sec-proofs}
In order to prove the main theorem, i.e.\ 
Theorem \ref{theorem-main-theorem}, we have to prove
Theorem \ref{theorem-continuity-of-svm-functional}
and Corollary \ref{cor-2-theorem-continuity-of-svm-functional}
at first.

\smallskip

\textbf{Proof of Theorem \ref{theorem-continuity-of-svm-functional}:}
Since the proof is somewhat involved, we start with a short outline.
The proof is divided into four parts. Part 1
is concerned with some important preparations.  We have to
show that $(f_{L^{\ast},\mathrm{P}_{n},\lambda})_{n\in\mathds{N}}$
converges to $f_{L^{\ast},\mathrm{P}_{0},\lambda}$ in $H$ if
the sequence of probability measures $(\mathrm{P}_{n})_{n\in\mathds{N}}$
weakly converges to the probability measure $\mathrm{P}_{0}$\,.
Let us now assume that there is a subsequence 
$(f_{L^{\ast},\mathrm{P}_{n_{\ell}},\lambda})_{\ell\in\mathds{N}}$
of $(f_{L^{\ast},\mathrm{P}_{n},\lambda})_{n\in\mathds{N}}$
which weakly converges to $f_{L^{\ast},\mathrm{P}_{0},\lambda}$ in $H$\,.
Then, it is shown in Part 2 and Part 3 that
\begin{eqnarray}  \label{outline-long-proof-of-main-theorem}
  \lim_{\ell\rightarrow\infty}
     \mathcal{R}_{L^{\ast},\mathrm{P}_{n_{\ell}}}
         (f_{L^{\ast},\mathrm{P}_{n_{\ell}},\lambda})
  &=&\mathcal{R}_{L^{\ast},\mathrm{P}_{0}}
         (f_{L^{\ast},\mathrm{P}_{0},\lambda})   
        \label{outline-long-proof-of-main-theorem-1} \\
  \lim_{\ell\rightarrow\infty}
     \mathcal{R}_{L^{\ast},\mathrm{P}_{n_{\ell}},\lambda}
         (f_{L^{\ast},\mathrm{P}_{n_{\ell}},\lambda})
  &=&\mathcal{R}_{L^{\ast},\mathrm{P}_{0},\lambda}
         (f_{L^{\ast},\mathrm{P}_{0},\lambda})\;\;.  
       \label{outline-long-proof-of-main-theorem-2} 
\end{eqnarray}
Because of
$$\|f\|_{H}^{2}
  \;=\;\frac{1}{\lambda}
         \Big(\mathcal{R}_{L^{\ast},\mathrm{P},\lambda}(f)
              -\mathcal{R}_{L^{\ast},\mathrm{P}}(f)
         \Big)
  \qquad\,\forall\,\mathrm{P}\in
       \mathcal{M}_{1}(\mathcal{X}\times\mathcal{Y})
   \quad\forall\,f\in H\;,
$$ 
it follows from (\ref{outline-long-proof-of-main-theorem-1}) 
and (\ref{outline-long-proof-of-main-theorem-2}) that
$\lim_{\ell\rightarrow\infty}
   \|f_{L^{\ast},\mathrm{P}_{n_{\ell}},\lambda}\|_{H}
 =\|f_{L^{\ast},\mathrm{P}_{0},\lambda}\|_{H}
$\,.
Since this convergence of the norms together with
weak convergence in the Hilbert space $H$ implies
(strong) convergence in $H$, we get that the subsequence
$(f_{L^{\ast},\mathrm{P}_{n_{\ell}},\lambda})_{\ell\in\mathds{N}}$
converges to $f_{L^{\ast},\mathrm{P}_{0},\lambda}$ in $H$\,.
Part 4 extends this result to the whole sequence  
$(f_{L^{\ast},\mathrm{P}_{n},\lambda})_{n\in\mathds{N}}$\,.
The main difficulty in the proof is the verification
of (\ref{outline-long-proof-of-main-theorem-1}) in Part 3.

In order to shorten notation, define
$$L_{f}^{\ast}:\;\,
    \mathcal{X}\times\mathcal{Y}\;\rightarrow\;\mathds{R}\,,
    \quad\; (x,y)\;\mapsto\;
  L^{\ast}\big(x,y,f(x)\big)=L(x,y,f(x))-L(x,y,0)
$$
for every measurable $f:\mathcal{X}\rightarrow\mathds{R}$\,.
Following e.g.\ \cite{vandervaart1998} and \cite{pollard2002},
we use the notation
$$\mathrm{P}g\;=\;\int g \,d\mathrm{P}
$$
for integrals of real-valued functions $g$ with respect to $\mathrm{P}$\,.
This leads to a very efficient notation which is
more intuitive here because, in the following, $\mathrm{P}$ 
rather acts as a linear
functional on a function space than as a probability measure on
a $\sigma$-algebra.

By use of these notations, we may write
$$\mathrm{P}L^{\ast}_{f}\;=\;
  \int L_{f}^{\ast}\,d\mathrm{P}\;=\;
  \mathcal{R}_{L^{\ast},\mathrm{P}}(f)
$$
for the (shifted) risk of $f\in H$\,. 
Accordingly, the (shifted) regularized risk of $f\in H$ is
$$\mathcal{R}_{L^{\ast},\mathrm{P},\lambda}(f)\;=\;
    \mathcal{R}_{L^{\ast},\mathrm{P}}(f)\,+\,\lambda\|f\|_{H}^2\;=\;
    \mathrm{P}L^{\ast}_{f}
    \,+\,\lambda\|f\|_{H}^2\;.
$$

  \textit{\underline{Part 1}}:
    Since the loss function $L$\,, 
    the shifted loss $L^{\ast}$ and the regularization
    parameter $\lambda\in(0,\infty)$ are fixed, we may drop
    them in the notation and write
    $$f_{\mathrm{P}}\;:=\;f_{L^{\ast},\mathrm{P},\lambda}
      \;=\;\SVM(\mathrm{P})
      \qquad\forall\,\mathrm{P}\in
          \mathcal{M}_{1}(\mathcal{X}\times\mathcal{Y})\;.
    $$
    Recall from Theorem \ref{theorem-L-star-trick-summary}
    that $f_{L^{\ast},\mathrm{P},\lambda}$ is equal to 
    the support vector machine $f_{L,\mathrm{P},\lambda}$
    if $f_{L,\mathrm{P},\lambda}$ exists. That is, we have
    $f_{\mathrm{P}}=f_{L,\mathrm{P},\lambda}$ in the latter case.
    According to \citet[(17),(16)]{christmann2009},
    \begin{eqnarray}
      \|f_{\mathrm{P}}\|_{\infty}&\leq&
      \frac{1}{\lambda}|L|_{1}\cdot\|k\|_{\infty}^{2}
            \label{theorem-continuity-of-svm-functional-1} \\
      \|f_{\mathrm{P}}\|_{H}&\leq&
      \sqrt{\frac{1}{\lambda}|L|_{1}\int |f_{\mathrm{P}}|\,d\mathrm{P}\,}
      \;\stackrel{(\ref{theorem-continuity-of-svm-functional-1})}{\leq}\;
      \frac{1}{\lambda}|L|_{1}\cdot\|k\|_{\infty}\;.
      \qquad \label{theorem-continuity-of-svm-functional-2}
    \end{eqnarray}
    for every $\mathrm{P}\in\mathcal{M}_{1}(\mathcal{X}\times\mathcal{Y})$\,.
    Since the kernel $k$ is continuous and bounded,
    \citet[Lemma 4.28]{steinwart2008} yields 
    \begin{eqnarray}\label{theorem-continuity-of-svm-functional-3}
      f\;\in\;\mathcal{C}_{b}(\mathcal{X})
      \qquad\forall\,f\in H\;.
    \end{eqnarray}
    Therefore, continuity of $L$ implies continuity of
    $$L_{f}^{\ast}\;:\;\;
      \mathcal{X}\times\mathcal{Y}\;\rightarrow\;\mathds{R}\,,\qquad
      (x,y)\;\mapsto\;
      L\big(x,y,f(x)\big)-L(x,y,0)
    $$
    for every $f\in H$\,.
    Furthermore, the uniform Lipschitz property of $L$ implies
    \begin{eqnarray*}
      \lefteqn{\sup_{x,y}\big|L_{f}^{\ast}(x,y)\big|
         \;=\;\sup_{x,y}\big|L(x,y,f(x))-L(x,y,0)\big| }\\
         &\leq&
              \sup_{x^{\prime},x,y}\big|L(x,y,f(x^{\prime}))-L(x,y,0)
                                   \big|\;\leq\;
              \sup_{x^{\prime}}|L|_{1}\cdot\big|f(x^{\prime})-0\big| 
         \;=\;|L|_{1}\|f\|_{\infty}
    \end{eqnarray*} 
    for every $f\in H$\,. 
    Hence, we obtain
    \begin{eqnarray}\label{theorem-continuity-of-svm-functional-4}
      L_{f}^{\ast}\;\in\;\mathcal{C}_{b}(\mathcal{X}\times\mathcal{Y})
      \qquad\forall\,f\in H\;.
    \end{eqnarray}
    In particular, the above calculation and 
    (\ref{theorem-continuity-of-svm-functional-1}) imply
    \begin{eqnarray}\label{theorem-continuity-of-svm-functional-5}
      \|L_{f_{\mathrm{P}}}^{\ast}\|_{\infty}\;\leq\;
      \frac{1}{\lambda}|L|_{1}^2\cdot\|k\|_{\infty}^2
      \quad\;\forall\,
       \mathrm{P}\in\mathcal{M}_{1}(\mathcal{X}\times\mathcal{Y})\;.
    \end{eqnarray}    
    
    For the remaining parts of the proof, let
    $\,(\mathrm{P}_{n})_{n\in\mathds{N}_{0}}\subset 
     \mathcal{M}_{1}(\mathcal{X}\times\mathcal{Y})
    $\, be any fixed sequence such that
    $$\mathrm{P}_{n}\;\longrightarrow\;
      \mathrm{P}_{0}
      \qquad(n\rightarrow\infty)
    $$
    in the weak topology on $\mathcal{M}_{1}(\mathcal{X}\times\mathcal{Y})$
    -- that is,
    \begin{eqnarray}\label{theorem-continuity-of-svm-functional-6}
      \lim_{n\rightarrow\infty}\mathrm{P}_{n}g\;=\;
          \mathrm{P}_{0}g
      \qquad\forall\,g\in\mathcal{C}_{b}(\mathcal{X}\times\mathcal{Y})\;.
    \end{eqnarray}
    In particular, (\ref{theorem-continuity-of-svm-functional-4}) and 
    (\ref{theorem-continuity-of-svm-functional-6}) imply
    \begin{eqnarray}\label{theorem-continuity-of-svm-functional-6a}
      \lim_{n\rightarrow\infty}\mathrm{P}_{n}L_{f}^{\ast}\;=\;
      \mathrm{P}_{0}L_{f}^{\ast}
      \qquad\forall\,f\in H\;.
    \end{eqnarray}
    In order to shorten the notation, define
    $$f_{n}\;:=\;f_{\mathrm{P}_{n}}
      \,=\,f_{L^{\ast},\mathrm{P}_{n},\lambda}\,=\,\SVM(\mathrm{P}_{n}) 
      \qquad\forall\,n\in\mathds{N}\cup\{0\}\;.
    $$
    Hence, we have to show that $(f_{n})_{n\in\mathds{N}}$ converges
    to $f_{0}$ in $H$ -- that is,
    \begin{eqnarray}\label{theorem-continuity-of-svm-functional-7}
      \lim_{n\rightarrow\infty}\|f_{n}-f_{0}\|_{H}
       \;=\;0\;.
    \end{eqnarray}
  \textit{\underline{Part 2}}: In this part of the proof,
    it is shown that
    \begin{eqnarray}\label{theorem-continuity-of-svm-functional-7a}
      \limsup_{n\rightarrow\infty}\,
         \mathrm{P}_{n}L^{\ast}_{f_{n}}+\lambda\|f_{n}\|^{2}_{H}
      \;\leq\;\mathrm{P}_{0}L^{\ast}_{f_{0}}+\lambda\|f_{0}\|^{2}_{H}\;.
    \end{eqnarray}
    Due to (\ref{theorem-continuity-of-svm-functional-4}), the 
    mapping
    $$\mathcal{M}_{1}(\mathcal{X}\times\mathcal{Y})\;\rightarrow\;
      \mathds{R}\,,\qquad
      \mathrm{P}\;\mapsto\;\mathrm{P}L^{\ast}_{f}+\lambda\|f\|^{2}_{H}
    $$
    is defined well and continuous for every $f\in H$\,.
    As being the (pointwise) infimum over a family of continuous
    functions, the function
    $$\mathcal{M}_{1}(\mathcal{X}\times\mathcal{Y})\;\rightarrow\;
      \mathds{R}\,,\qquad\mathrm{P}\;\mapsto\;
        \inf_{f\in H}\,
          \big(\mathrm{P}L^{\ast}_{f}+\lambda\|f\|^{2}_{H}\big)
    $$
    is upper semicontinuous;
    see, e.g., \citet[Prop.\ 1.1.36]{denkowski2003}.
    Therefore, the definition of $f_{n}$ implies
    \begin{eqnarray*}
      \lefteqn{\limsup_{n\rightarrow\infty}\,
         \big(\mathrm{P}_{n}L^{\ast}_{f_{n}}+\lambda\|f_{n}\|^{2}_{H}
         \big)\;=\;
      \limsup_{n\rightarrow\infty}\,
         \inf_{f\in H}\,
            \big(\mathrm{P}_{n}L^{\ast}_{f}+\lambda\|f\|^{2}_{H}\big)
          \;\leq}\\
      &&\leq\; 
         \inf_{f\in H}\,
           \big(\mathrm{P}_{0}L^{\ast}_{f}+\lambda\|f\|^{2}_{H}\big)
        \;=\;\mathrm{P}_{0}L^{\ast}_{f_{0}}+\lambda\|f_{0}\|^{2}_{H}\;\;.
        \qquad\qquad\qquad
    \end{eqnarray*}

  \textit{\underline{Part 3}}: In this part of the proof,
    the following statement is shown:
    
    Let $(f_{n_{\ell}})_{\ell\in\mathds{N}}$ be a subsequence of
    $(f_{n})_{n\in\mathds{N}}$ 
    and assume that $(f_{n_{\ell}})_{\ell\in\mathds{N}}$ 
    converges weakly in
    $H$ to some $f_{0}^{\prime}\in H$.
    Then, the following three assertions are true:
    \begin{eqnarray}
      &&
        \lim_{\ell\rightarrow\infty}
        \mathrm{P}_{n_{\ell}}L^{\ast}_{f_{n_{\ell}}}
        \;=\;
        \mathrm{P}_{0}L^{\ast}_{f_{0}^{\prime}} \qquad
        \label{theorem-continuity-of-svm-functional-8} \\
      &&\qquad\quad
        f_{0}^{\prime}\;=\;f_{0} \qquad
        \label{theorem-continuity-of-svm-functional-9} \\
      &&
        \lim_{\ell\rightarrow\infty}
        \|f_{n_{\ell}}-f_{0}\|_{H}
        \;=\;0 \;\;.\quad
        \label{theorem-continuity-of-svm-functional-10} 
    \end{eqnarray}
    In order to prove this, we will also have to deal with
    subsequences of the subsequence $(f_{n_{\ell}})_{\ell\in\mathds{N}}$\,. 
    As this would lead to a 
    somewhat cumbersome notation, we define
    $$\mathrm{P}^{\prime}_{\ell}\;:=\;\mathrm{P}_{n_{\ell}}
        \quad\text{and}\quad
      f^{\prime}_{\ell}\;:=\;f_{n_{\ell}}
      \qquad \ell\in\mathds{N}\;.
    $$
    Thus, 
    $f^{\prime}_{\ell}=f_{L^{\ast},\mathrm{P}_{n_{\ell}},\lambda}$
    for every $\ell\in\mathds{N}$\,.
    Then, the assumption of weak convergence in 
    the Hilbert space $H$ equals 
    \begin{eqnarray}\label{theorem-continuity-of-svm-functional-11}
      \lim_{\ell\rightarrow\infty}
      \langle f^{\prime}_{\ell},h\rangle_{H}
      \;\;=\;\; 
      \langle f^{\prime}_{0},h \rangle_{H}
      \qquad\forall\,h\in H\;.
    \end{eqnarray}
    First of all, we show 
    (\ref{theorem-continuity-of-svm-functional-8}) 
    by proving
    \begin{eqnarray}\label{theorem-continuity-of-svm-functional-12}
      \limsup_{\ell\rightarrow\infty}
      \big|\mathrm{P}_{\ell}^{\prime}L^{\ast}_{f_{\ell}^{\prime}}-
           \mathrm{P}_{0}L^{\ast}_{f_{0}^{\prime}}\big|
      \;\leq\;\varepsilon_{0} 
    \end{eqnarray}
    for every fixed $\varepsilon_{0}>0$. In order to do this, 
    fix any $\varepsilon_{0}>0$ and define
    \begin{eqnarray}\label{theorem-continuity-of-svm-functional-13}
      \varepsilon\;:=\;
      \frac{\varepsilon_{0}}%
           {\,|L|_{1}\cdot
            \big({\textstyle\frac{1}{\lambda}}|L|_{1}\cdot\|k\|_{\infty}^2
                 +\|f_{0}^{\prime}\|_{\infty}
            \big)
           }
      \;\;>\;\;0\;.
    \end{eqnarray}   
    The following calculation shows that the 
    sequence of functions 
    $(f_{\ell}^{\prime})_{\ell\in\mathds{N}}$ is uniformly continuous on
    $\mathcal{X}$\,.
    For any convergent sequence $x_{m}\rightarrow x_{0}$ in
    $\mathcal{X}$\,, we have
    \begin{eqnarray*}
      \lefteqn{\limsup_{m\rightarrow\infty} \sup_{\ell\in\mathds{N}}\,
          \big|f_{\ell}^{\prime}(x_{m})-f_{\ell}^{\prime}(x_{0})\big| }\\
      &&\;\;=\;\limsup_{m\rightarrow\infty}\, \sup_{\ell\in\mathds{N}}\,
                \big|\langle f_{\ell}^{\prime},\Phi(x_{m})\rangle_{H}-
                     \langle f_{\ell}^{\prime},\Phi(x_{0})\rangle_{H}
                \big| \\
      &&\;\;=\;\limsup_{m\rightarrow\infty}\, \sup_{\ell\in\mathds{N}}\,
            \big|\langle f_{\ell}^{\prime},\Phi(x_{m})-\Phi(x_{0})\rangle_{H}
            \big| \\
      &&\;\;\leq\;\limsup_{m\rightarrow\infty}\, \sup_{\ell\in\mathds{N}}\,
                \|f_{\ell}^{\prime}\|_{H}\cdot
                     \|\Phi(x_{m})-\Phi(x_{0})\|_{H}\\
      &&\;\stackrel{(\ref{theorem-continuity-of-svm-functional-2})}{\leq}\; 
             \frac{1}{\lambda}\,|L|_{1}\cdot\|k\|_{\infty}\cdot
               \limsup_{m\rightarrow\infty}\,
                     \|\Phi(x_{m})-\Phi(x_{0})\|_{H}
         \;=\;0\qquad\quad
    \end{eqnarray*} 
    where the first equality follows from the properties of the
    RKHS $H$ and 
    the last equality follows from 
    \citet[Lemma 4.29]{steinwart2008}.

    Since
    $\mathcal{X}\times\mathcal{Y}$ is
    a Polish space, weak convergence of
    $(\mathrm{P}_{\ell}^{\prime})_{\ell\in\mathds{N}}$ implies 
    uniform tightness of $(\mathrm{P}_{\ell}^{\prime})_{\ell\in\mathds{N}}$
    (see e.g.\ \citet[Theorem 11.5.3]{dudley1989}).
    That is, there is a
    compact subset $K_{\varepsilon}\subset\mathcal{X}\times\mathcal{Y}$
    such that
    \begin{eqnarray}\label{theorem-continuity-of-svm-functional-14}
      \limsup_{\ell\rightarrow\infty}\,\mathrm{P}_{\ell}^{\prime}
      \big(K_{\varepsilon}^{\,\textup{c}}\big)
      \;<\;\varepsilon\;.
    \end{eqnarray}
    Since $K_{\varepsilon}$ is compact and the projection 
    $$\tau_{\mathcal{X}}:\;\;\mathcal{X}\times\mathcal{Y}\;\rightarrow\;
      \mathcal{X}\,,\qquad
      (x,y)\;\mapsto\;x
    $$
    is continuous, 
    $\,\tilde{K}_{\varepsilon}:=\tau_{X}(K_{\varepsilon})\,$ is
    compact in $\mathcal{X}$\,. For every $\ell\in\mathds{N}_{0}$\,,
    the restriction of $f_{\ell}^{\prime}$ on
    $\tilde{K}_{\varepsilon}$ is denoted by $\tilde{f}_{\ell}^{\prime}$\,.
    As the sequence 
    $(f_{\ell}^{\prime})_{\ell\in\mathds{N}}$ is uniformly continuous
    on $\mathcal{X}$ 
    and uniformly bounded 
    in $\mathcal{C}_{b}(\mathcal{X})$
    (see (\ref{theorem-continuity-of-svm-functional-1})), 
    the sequence of the restrictions 
    $(\tilde{f}_{\ell}^{\prime})_{\ell\in\mathds{N}}$
    has the corresponding properties on $\tilde{K}_{\varepsilon}$\,.
    That is, $(\tilde{f}_{\ell}^{\prime})_{\ell\in\mathds{N}}$
    is uniformly continuous on $\tilde{K}_{\varepsilon}$ 
    and uniformly bounded 
    in $\mathcal{C}_{b}(\tilde{K}_{\varepsilon})$\,.
    Hence, the Arzela-Ascoli-Theorem 
    -- see \citet[Theorem VI.3.8]{conway1985} -- assures that
    $(\tilde{f}_{\ell}^{\prime})_{\ell\in\mathds{N}}$
    is totally bounded and, therefore, relatively compact
    in $\mathcal{C}_{b}(\tilde{K}_{\varepsilon})$
    (since $\mathcal{C}_{b}(\tilde{K}_{\varepsilon})$ is a complete metric
    space);
    see e.g.\ \citet[Theorem I.6.15]{dunford1958}.
    
    The following reasoning shows that
    $(\tilde{f}_{\ell}^{\prime})_{\ell\in\mathds{N}}$ converges to
    $\tilde{f}_{0}^{\prime}$ in 
    $\mathcal{C}_{b}(\tilde{K}_{\varepsilon})$\,, i.e.\
    \begin{eqnarray}\label{theorem-continuity-of-svm-functional-15}
      \lim_{\ell\rightarrow\infty}\;
      \sup_{x\in \tilde{K}_{\varepsilon}}
        \big|f_{\ell}^{\prime}(x)-f_{0}^{\prime}(x)\big|
        \;=\;0\;.
    \end{eqnarray}
    We will show (\ref{theorem-continuity-of-svm-functional-15}) 
    by contradiction.
    If (\ref{theorem-continuity-of-svm-functional-15}) is not true,
    then there is a $\delta>0$ and a subsequence 
    $(\tilde{f}_{\ell_{j}}^{\prime})_{j\in\mathds{N}}$ such that
    \begin{eqnarray}\label{theorem-continuity-of-svm-functional-15b}
      \sup_{x\in \tilde{K}_{\varepsilon}}
        \big|f_{\ell_{j}}^{\prime}(x)-f_{0}^{\prime}(x)\big|
        \;>\;\delta
      \qquad\forall\,j\in\mathds{N}\;.
    \end{eqnarray}
    Relative compactness of
    $(\tilde{f}_{\ell}^{\prime})_{\ell\in\mathds{N}}$
    implies that 
    there is a further subsequence
    $(\tilde{f}_{\ell_{j_{m}}}^{\prime})_{m\in\mathds{N}}$
    which converges in $\mathcal{C}_{b}(\tilde{K}_{\varepsilon})$ to 
    some
    $\tilde{h}_{0}\in\mathcal{C}_{b}(\tilde{K}_{\varepsilon})$\,.
    Then,
    \begin{eqnarray*}
      \tilde{h}_{0}(x)
      &=& \lim_{m\rightarrow\infty}\tilde{f}_{\ell_{j_{m}}}^{\prime}(x)
         \;=\;\lim_{m\rightarrow\infty}f_{\ell_{j_{m}}}^{\prime}(x)
         \;=\;\lim_{m\rightarrow\infty}
                \langle f_{\ell_{j_{m}}}^{\prime},\Phi(x)\rangle_{H}\;=\\
      &\stackrel{(\ref{theorem-continuity-of-svm-functional-11})}{=}&
             \langle f_{0}^{\prime},\Phi(x)\rangle_{H}
          \;=\;f_{0}^{\prime}(x)
          \;=\;\tilde{f}_{0}^{\prime}(x)\;\;.
    \end{eqnarray*}
    for every $x\in \tilde{K}_{\varepsilon}$\,.
    That is, $\tilde{f}_{0}^{\prime}$ is the limit of
    $(\tilde{f}_{\ell_{j_{m}}}^{\prime})_{m\in\mathds{N}}$ 
    -- which is the desired contradiction to
    (\ref{theorem-continuity-of-svm-functional-15b}).
    Therefore, (\ref{theorem-continuity-of-svm-functional-15})
    is true.
 
    Now, we can prove
   (\ref{theorem-continuity-of-svm-functional-12}):  
    Firstly, the triangle inequality and the Lipschitz
    continuity of $L$ yield
    \begin{eqnarray*}
      \lefteqn{\limsup_{\ell\rightarrow\infty}
           \big|\mathrm{P}_{\ell}^{\prime}L^{\ast}_{f_{\ell}^{\prime}}-
                \mathrm{P}_{0}L^{\ast}_{f_{0}^{\prime}}
           \big|
           \;\leq\;\limsup_{\ell\rightarrow\infty}\,\,
           \big|\mathrm{P}_{\ell}^{\prime}L^{\ast}_{f_{\ell}^{\prime}}-
                \mathrm{P}_{\ell}^{\prime}L^{\ast}_{f_{0}^{\prime}}
           \big|+
           \big|\mathrm{P}_{\ell}^{\prime}L^{\ast}_{f_{0}^{\prime}}-
                \mathrm{P}_{0}L^{\ast}_{f_{0}^{\prime}}
           \big| }\\
      &\stackrel{(\ref{theorem-continuity-of-svm-functional-6a})}{=}&
          \limsup_{\ell\rightarrow\infty}
           \big|\mathrm{P}_{\ell}^{\prime}L^{\ast}_{f_{\ell}^{\prime}}-
                \mathrm{P}_{\ell}^{\prime}L^{\ast}_{f_{0}^{\prime}}
           \big| \\
      &=&\limsup_{\ell\rightarrow\infty}
           \left|\int \! L(x,y,f_{\ell}^{\prime}(x))-L(x,y,f_{0}^{\prime}(x))
                 \,\,d\mathrm{P}_{\ell}^{\prime}
           \right| \\ 
      &\leq&\limsup_{\ell\rightarrow\infty}
           \int |L|_{1}\cdot\big|f_{\ell}^{\prime}(x)-f_{0}^{\prime}(x)\big|
           \,\,\mathrm{P}_{\ell}^{\prime}(d(x,y))\;=\\
      &=&|L|_{1}\cdot\limsup_{\ell\rightarrow\infty}\,\,
              \Bigg(\int_{K_{\varepsilon}}
                \big|f_{\ell}^{\prime}(x)-f_{0}^{\prime}(x)\big|
              \,\,\mathrm{P}_{\ell}^{\prime}(d(x,y))
              \,+\\
         &&\qquad\qquad\qquad\qquad\qquad +\;
              \int_{K_{\varepsilon}^{\textup{c}}}
                \big|f_{\ell}^{\prime}(x)-f_{0}^{\prime}(x)\big|
              \,\,\mathrm{P}_{\ell}^{\prime}(d(x,y)) \Bigg)
          \;\;\;.\quad\quad \\
    \end{eqnarray*}
    Secondly, using $\,\tilde{K}_{\varepsilon}=\tau_{X}(K_{\varepsilon})\,$,
    we obtain
    \begin{eqnarray*}
      \lefteqn{
        \limsup_{\ell\rightarrow\infty}\,\,
              \int_{K_{\varepsilon}}
                \big|f_{\ell}^{\prime}(x)-f_{0}^{\prime}(x)\big|
              \,\,\mathrm{P}_{\ell}^{\prime}(d(x,y)) } \\
      &\leq&\limsup_{\ell\rightarrow\infty}\,
              \sup_{(x,y)\in K_{\varepsilon}}
               \big|f_{\ell}^{\prime}(x)-f_{0}^{\prime}(x)\big|
        \,=\,
            \limsup_{\ell\rightarrow\infty}\,
              \sup_{x\in\tilde{K}_{\varepsilon}}
               \big|f_{\ell}^{\prime}(x)-f_{0}^{\prime}(x)\big|
        \;\stackrel{(\ref{theorem-continuity-of-svm-functional-15})}{=}\;
            0\,.
    \end{eqnarray*}
    Thirdly,
    \begin{eqnarray*}
      \lefteqn{
        \limsup_{\ell\rightarrow\infty}\,\,
        \int_{K_{\varepsilon}^{\textup{c}}}
                \big|f_{\ell}^{\prime}(x)-f_{0}^{\prime}(x)\big|
              \,\,\mathrm{P}_{\ell}^{\prime}(d(x,y))}\\
      &\leq&\limsup_{\ell\rightarrow\infty}\,\,
           \mathrm{P}_{\ell}^{\prime}
               \big(K_{\varepsilon}^{\textup{c}}\big)
               \cdot\big(\|f_{\ell}^{\prime}\|_{\infty}
                         +\|f_{0}^{\prime}\|_{\infty}
                    \big) \\
      &\stackrel{(\ref{theorem-continuity-of-svm-functional-14})}{\leq}&
            \limsup_{\ell\rightarrow\infty}\,\,\varepsilon
                    \cdot\big(\|f_{\ell}^{\prime}\|_{\infty}
                              +\|f_{0}^{\prime}\|_{\infty}
                         \big)
        \;\;\stackrel{(\ref{theorem-continuity-of-svm-functional-1}),
                    (\ref{theorem-continuity-of-svm-functional-13})
                   }{=}\;\;
           \frac{\varepsilon_{0}}{\,|L|_{1}}\;\;\;.
    \end{eqnarray*}
    Combining these three calculations
    proves (\ref{theorem-continuity-of-svm-functional-12}).
    Since $\varepsilon_{0}>0$ was arbitrarily chosen in
    (\ref{theorem-continuity-of-svm-functional-12}),
    this proves
    (\ref{theorem-continuity-of-svm-functional-8}).
    
    Next, we prove (\ref{theorem-continuity-of-svm-functional-9}):   
    Due to weak convergence of $(f_{n_{\ell}})_{\ell\in\mathds{N}}$ 
    in $H$, 
    it follows from 
    \citet[Exercise V.1.9]{conway1985} that
    \begin{eqnarray}\label{theorem-continuity-of-svm-functional-16}
      \|f_{0}^{\prime}\|_{H}
      \;\leq\;
      \liminf_{\ell\rightarrow\infty} \|f_{n_{\ell}}\|_{H}\;.
    \end{eqnarray}
    Therefore, the definition of $f_{0}=f_{L^{\ast},\mathrm{P}_{0},\lambda}$
    implies
    \begin{eqnarray*}
      \lefteqn{\mathrm{P}_{0}L^{\ast}_{f_{0}}+\lambda\|f_{0}\|^{2}_{H}
         \;=\;\inf_{f\in H}\,
                \mathrm{P}_{0}L^{\ast}_{f}+\lambda\|f\|^{2}_{H} }\\
      &\leq&\mathrm{P}_{0}L^{\ast}_{f_{0}^{\prime}}
              +\lambda\|f_{0}^{\prime}\|^{2}_{H}
        \;\stackrel{(\ref{theorem-continuity-of-svm-functional-8}),
                    (\ref{theorem-continuity-of-svm-functional-16})}%
                   {\leq}\;
             \liminf_{\ell\rightarrow\infty}\,
                \mathrm{P}_{n_{\ell}}L^{\ast}_{f_{n_{\ell}}}
                 +\lambda\|f_{n_{\ell}}\|^{2}_{H} \\
      &\leq&\limsup_{\ell\rightarrow\infty}\,
                \mathrm{P}_{n_{\ell}}L^{\ast}_{f_{n_{\ell}}}
                 +\lambda\|f_{n_{\ell}}\|^{2}_{H}
        \;\stackrel{(\ref{theorem-continuity-of-svm-functional-7a})}%
                   {\leq}\;
             \mathrm{P}_{0}L^{\ast}_{f_{0}}+\lambda\|f_{0}\|^{2}_{H}\;\;.
    \end{eqnarray*} 
    Due to this calculation, it follows that
    \begin{eqnarray}\label{theorem-continuity-of-svm-functional-17}
      \mathrm{P}_{0}L^{\ast}_{f_{0}}+\lambda\|f_{0}\|^{2}_{H}
      \;=\;\inf_{f\in H}\,
                \mathrm{P}_{0}L^{\ast}_{f}+\lambda\|f\|^{2}_{H}
      \;=\;\mathrm{P}_{0}L^{\ast}_{f_{0}^{\prime}}
              +\lambda\|f_{0}^{\prime}\|^{2}_{H}
    \end{eqnarray}
    and
    \begin{eqnarray}\label{theorem-continuity-of-svm-functional-18}
      \mathrm{P}_{0}L^{\ast}_{f_{0}}+\lambda\|f_{0}\|^{2}_{H}
      \;=\;\lim_{\ell\rightarrow\infty}\,
                \mathrm{P}_{n_{\ell}}L^{\ast}_{f_{n_{\ell}}}
                 +\lambda\|f_{n_{\ell}}\|^{2}_{H}\;.
    \end{eqnarray}
    According to Theorem \ref{theorem-L-star-trick-summary},
    $f_{0}=f_{L^{\ast},\mathrm{P}_{0},\lambda}$ is the unique minimizer of
    the function
    $$H\;\rightarrow\;\mathds{R}\,,\quad
      f\;\mapsto\;\mathrm{P}_{0}L^{\ast}_{f}+\lambda\|f\|^{2}_{H}
    $$ 
    and, therefore, (\ref{theorem-continuity-of-svm-functional-17})
    implies $f_{0}=f_{0}^{\prime}$ \,--\,
    i.e. (\ref{theorem-continuity-of-svm-functional-9}).
    
    Completing Part 3 of the proof, 
    (\ref{theorem-continuity-of-svm-functional-10}) is shown now:
    \begin{eqnarray*}
      \lim_{\ell\rightarrow\infty} \|f_{n_{\ell}}\|_{H}^2
      &=&\lim_{\ell\rightarrow\infty}\,\frac{1}{\lambda}
         \Big(\big(\mathrm{P}_{n_{\ell}}L^{\ast}_{f_{n_{\ell}}}
                 +\lambda\|f_{n_{\ell}}\|^{2}_{H}\big)
              \,\,-\,\,\mathrm{P}_{n_{\ell}}L^{\ast}_{f_{n_{\ell}}}
         \Big)\\
      &\stackrel{(\ref{theorem-continuity-of-svm-functional-8}),%
                 (\ref{theorem-continuity-of-svm-functional-18})}{=}&
         \,\frac{1}{\lambda}
         \Big(\big(\mathrm{P}_{0}L^{\ast}_{f_{0}}
                 +\lambda\|f_{0}\|^{2}_{H}\big)
              \,\,-\,\,\mathrm{P}_{0}L^{\ast}_{f_{0}}
         \Big)
         \;=\;\|f_{0}\|_{H}^2\;\;.
    \end{eqnarray*}
    By assumption, the sequence 
    $(f_{n_{\ell}})_{\ell\in\mathds{N}}$ converges weakly to
    some $f_{0}^{\prime}\in H$ and by 
    (\ref{theorem-continuity-of-svm-functional-9}),
    we know that $f_{0}^{\prime}=f_{0}$.
    In addition, we have proven 
    $\lim_{\ell\rightarrow\infty} \|f_{n_{\ell}}\|_{H}=\|f_{0}\|_{H}$
    now. This 
    convergence of the norms together with weak convergence 
    implies strong convergence in the Hilbert space $H$,
    -- see, e.g., \citet[Exercise V.1.8]{conway1985}.
    That is, we have proven
    (\ref{theorem-continuity-of-svm-functional-10}).
    
  \textit{\underline{Part 4}}: In this final part of the proof,
    (\ref{theorem-continuity-of-svm-functional-7}) is shown.
    This is done by contradiction:
    If (\ref{theorem-continuity-of-svm-functional-7}) is not true,
    there is an $\varepsilon>0$ and a subsequence 
    $(f_{n_{\ell}})_{\ell\in\mathds{N}}$ of $(f_{n})_{n\in\mathds{N}}$ 
    such that
    \begin{eqnarray}\label{theorem-continuity-of-svm-functional-19}
      \|f_{n_{\ell}}-f_{0}\|_{H}\;>\;\varepsilon
      \qquad\forall\,\ell\in\mathds{N}
    \end{eqnarray}
    According to (\ref{theorem-continuity-of-svm-functional-2})\,,
    $(f_{n_{\ell}})_{\ell\in\mathds{N}}
     =(f_{\mathrm{P}_{n_{\ell}}})_{\ell\in\mathds{N}}
    $
    is bounded in $H$\,. Hence, the sequence
    $(f_{n_{\ell}})_{\ell\in\mathds{N}}$ contains a further subsequence
    that weakly converges in $H$ to some $f_{0}^{\prime}$\,;
    see e.g. \citet[Corollary IV.4.7]{dunford1958}. 
    Without loss of generality,
    we may therefore assume that $(f_{n_{\ell}})_{\ell\in\mathds{N}}$ 
    weakly converges in $H$ to some $f_{0}^{\prime}$\,.
    (Otherwise, we can choose another subsequence in  
    (\ref{theorem-continuity-of-svm-functional-19})).
    Next, it follows from Part 3, that $(f_{n_{\ell}})_{\ell\in\mathds{N}}$ 
    strongly converges in $H$ to $f_{0}$ --
    which is a contradiction to 
    (\ref{theorem-continuity-of-svm-functional-19}).
\hfill$\Box$

\medskip

\textbf{Proof of Corollary 
     \ref{cor-2-theorem-continuity-of-svm-functional}:}
  Let $(D_{n,m})_{m\in\mathds{N}}$ be a sequence in 
  $(\mathcal{X}\times\mathcal{Y})^n$ which converges to
  some $D_{n,0}\in(\mathcal{X}\times\mathcal{Y})^n$\,.
  Then, the corresponding sequence of empirical measures
  $\big(\mathds{P}_{D_{n,m}}\big)_{m\in\mathds{N}}$ weakly converges
  in $\mathcal{M}_{1}(\mathcal{X}\times\mathcal{Y})$ to
  $\mathds{P}_{D_{n,0}}$\,. Therefore, the statement follows from
  Theorem \ref{theorem-continuity-of-svm-functional} 
  and (\ref{representation-of-empirical-svm}).
\hfill$\Box$

\medskip

Based on \cite{cuevas1988}, 
the main theorem  
essentially is a consequence of 
Theorem \ref{theorem-continuity-of-svm-functional}.

\smallskip

\textbf{Proof of Theorem \ref{theorem-main-theorem}:}
  According to Corollary 
  \ref{cor-2-theorem-continuity-of-svm-functional},
  the SVM-estimator 
  $$\SVM_{n}\;:\;\;(\mathcal{X}\times\mathcal{Y})^n\;\rightarrow\;H\,,
    \qquad
    D_{n}\;\mapsto\;
    f_{L,D_{n},\lambda}
  $$
  is continuous and, therefore,
  measurable with respect to the 
  Borel-$\sigma$-algebras
  for every $n\in\mathds{N}$\,. The mapping
  $$\SVM\;:\;\;\mathcal{M}_{1}(\mathcal{X}\times\mathcal{Y})
    \;\rightarrow\;H\,,\quad\;
    \mathrm{P}\;\mapsto\;f_{L^{\ast},\mathrm{P},\lambda}
  $$
  is a continuous functional 
  due to Theorem \ref{theorem-continuity-of-svm-functional}.
  Furthermore,
  $$\SVM_{n}(D_{n})\;=\;
    \SVM\big(\mathds{P}_{D_{n}}\big)
    \qquad\forall\,D_{n}\in(\mathcal{X}\times\mathcal{Y})^n
    \qquad\quad\forall\,n\in\mathds{N}\;.
  $$  
  As already mentioned in
  Section \ref{subsec-setup}, $H$ is a separable Hilbert space
  and, therefore, a Polish space.
  Hence, the sequence of SVM-estimators 
  $(\SVM_{n})_{n\in\mathds{N}}$ is qualitatively robust
  according to \citet[Theorem 2]{cuevas1988}.
\hfill$\Box$

\medskip

\textbf{Proof of Corollary \ref{cor-consistency}:}
  Let $\mathds{P}_{\datenzfv_{n}}$ denote the function which
  maps $\omega\in\Omega$ to the empirical measure
  $\frac{1}{n}\sum_{i=1}^{n}\delta_{(X_{i}(\omega),Y_{i}(\omega))}$\,.
  According to Varadarajan's Theorem 
  (\citet[Theorem 11.4.1]{dudley1989}), there is a 
  set $N\in\mathcal{A}$ such that
  $\mathrm{Q}(N)=0$ and
  $\mathds{P}_{\datenzfv_{n}(\omega)}$ weakly converges to
  $\mathrm{P}$ for every $\omega\in\Omega\setminus N$\,.
  Then, Theorem \ref{theorem-continuity-of-svm-functional}
  implies
  $$\lim_{n\rightarrow\infty}
          \|f_{L^{\ast},\datenzfv_{n}(\omega),\lambda}
                 -f_{L^{\ast},\mathrm{P},\lambda}\|_{H}
    \;\stackrel{(\ref{representation-of-empirical-svm})}{=}\;
    \lim_{n\rightarrow\infty}
          \|\SVM(\mathds{P}_{\datenzfv_{n}(\omega)})-\SVM(\mathrm{P})\|_{H}
    \;=\;0
  $$
  for every $\omega\in\Omega\setminus N$\,. This proves (a)
  and, due to \citet[Lemma 4.28]{steinwart2008}, (b).
  The Lipschitz continuity of $L^{\ast}$ implies
  \begin{eqnarray*}
    \lefteqn{
      \big|\mathcal{R}_{L^{\ast},\mathrm{P}}
              (f_{L^{\ast},\datenzfv_{n}(\omega),\lambda})
          -\mathcal{R}_{L^{\ast},\mathrm{P}}(f_{L^{\ast},\mathrm{P},\lambda})
      \big| }\\
    &&=\;\left|\int L(x,y,f_{L^{\ast},\datenzfv_{n}(\omega),\lambda}(x))
                     -L(x,y,f_{L^{\ast},\mathrm{P},\lambda}(x))
                 \,\mathrm{P}(d(x,y))
           \right| \\
    &&\leq\;\int 
        \sup_{x^{\prime},y^{\prime}} 
        \big|L(x^{\prime},y^{\prime},
                       f_{L^{\ast},\datenzfv_{n}(\omega),\lambda}(x))
             -L(x^{\prime},y^{\prime},
                       f_{L^{\ast},\mathrm{P},\lambda}(x))
        \big|
        \,\mathrm{P}(d(x,y)) \\
    &&\leq\;\int |L|_{1} \cdot
        \big|f_{L^{\ast},\datenzfv_{n}(\omega),\lambda}(x)
             -f_{L^{\ast},\mathrm{P},\lambda}(x)
        \big|
        \,\mathrm{P}(d(x,y)) \\
    &&\leq\;|L|_{1} \cdot
        \big\|f_{L^{\ast},\datenzfv_{n}(\omega),\lambda}
             -f_{L^{\ast},\mathrm{P},\lambda}
        \big\|_{\infty}
  \end{eqnarray*}
  for every $\omega\in\Omega$\,. According to (b), the last term converges to
  0 for $\mathrm{Q}$\,-\,almost every $\omega\in\Omega$ and this implies
  (d). Finally, (c) follows from (a) and (d).
   
  If $f_{L,\mathrm{P},\lambda}$ exists,
  then $f_{L^{\ast},\mathrm{P},\lambda}$ is equal to
  $f_{L,\mathrm{P},\lambda}$ 
  (Theorem \ref{theorem-L-star-trick-summary}).
  In particular, there is an $f\in H$ such that
  $(x,y)\mapsto L(x,y,f(x))$ is $\mathrm{P}$\,-\,integrable.
  Since Lipschitz-continuity of $L$ and 
  $H\subset\mathcal{C}_{b}(\mathcal{X})$ (see
  \citet[Lemma 4.28]{steinwart2008}) implies
  $\mathrm{P}$\,-\,integrability of
  $(x,y)\mapsto L^{\ast}(x,y,f(x))=L(x,y,f(x))-L(x,y,0)$\,,
  we get that $(x,y)\mapsto L(x,y,0)$ is also $\mathrm{P}$\,-\,integrable.
  Therefore, $\mathcal{R}_{L^{\ast},\mathrm{P}}(f)$ is equal to
  $\mathcal{R}_{L,\mathrm{P}}(f)-\mathcal{R}_{L,\mathrm{P}}(0)$
  for every $f\in H$,
  and $\mathcal{R}_{L,\mathrm{P}}(0)$ is a finite constant which does not
  depend on $f$\,.
  Furthermore, 
  $f_{L^{\ast},D_{n},\lambda}
   =f_{L,D_{n},\lambda}
  $
  for every $D_{n}\in(\mathcal{X}\times\mathcal{Y})^{n}$\,; 
  see Section \ref{subsec-setup}.
  Hence, the original assertions (a)--(d)
  for $L^{\ast}$ turn into
  the corresponding assertions
  for $L$ instead of $L^{\ast}$\,.
\hfill$\Box$

\medskip

\textbf{Proof of Theorem \ref{theorem-continuity-bochner}:} 
  If $\Psi=0$\,, the statement is true.
  Assume $\Psi\not=0$ now and assume that the statement 
  of the theorem is
  not true. Then, there is an $\varepsilon>0$ and
  a subsequence $(\mathrm{P}_{n_\ell})_{\ell\in\mathds{N}}$
  such that
  \begin{eqnarray}\label{theorem-continuity-bochner-1001}
    \bigg\|\int \Psi\,d\mathrm{P}_{n_l}-
           \int \Psi\,d\mathrm{P}_{0}
    \bigg\|_H
    \;>\;\varepsilon
    \qquad\forall\,\ell\in\mathds{N}\;.
  \end{eqnarray} 
  Since the sequence
  $(\mathrm{P}_{n})_{n\in\mathds{N}}$ weakly converges
  to $\mathrm{P}_0$, 
  it is uniformly tight; see, e.g., \citep[Theorem 11.5.3]{dudley1989}. 
  That is, there is a 
  compact subset $K\subset\mathcal{Z}$ such that
  \begin{eqnarray}\label{theorem-continuity-bochner-1}
    \mathrm{P}_{n_\ell}\big(\mathcal{Z}\setminus K\big)\;<\;
    \frac{\varepsilon}{4\sup_{z}\|\Psi(z)\|_{H}}
    \qquad\forall\,\ell\in\mathds{N}_{0}\;.
  \end{eqnarray} 
  For every $\ell\in\mathds{N}$\,, let 
  $\tilde{\mathrm{P}}_{n_\ell}$
  denote the restriction of $\mathrm{P}_{n_\ell}$ to the 
  Borel-$\sigma$-algebra
  $\mathfrak{B}(K)$ of $K$\,. 
  Let $\tilde{\Psi}$ denote the restriction of $\Psi$ to $K$.
  Since
  $K$ is a compact Polish space, the set $\mathcal{M}(K)$
  of all finite signed measures on $\mathfrak{B}(K)$ is the 
  dual space of $\mathcal{C}(K)$ (the set of
  all continuous functions $f:K\rightarrow\mathds{R}$); see
  e.g.\ \citep[Theorem 7.1.1 and 7.4.1]{dudley1989}. 
  Accordingly, $\mathcal{M}(K)$ is precisely the set of all 
  (real) measures
  in the sense of \citep[Section III.1]{bourbaki2004integration};
  see also 
  \citep[Subsection III.1.5 and III.1.8]{bourbaki2004integration}.
  Since $(\tilde{\mathrm{P}}_{n_\ell})_{\ell\in\mathds{N}}$ 
  is relatively compact in the vague topology of $\mathcal{M}(K)$
  \citep[Subsection III.1.9]{bourbaki2004integration},
  we may assume without loss of generality that
  $(\tilde{\mathrm{P}}_{n_\ell})_{\ell\in\mathds{N}}$
  vaguely converges to some positive finite measure 
  $\tilde{\mathrm{P}}_0^\prime$. (Otherwise,
  we may replace $(\tilde{\mathrm{P}}_{n_\ell})_{\ell\in\mathds{N}}$ 
  by a further subsequence.)
  According to \citep[p.\ III.40]{bourbaki2004integration},
  vague convergence implies
  \begin{eqnarray}\label{theorem-continuity-bochner-1002}
    \int \tilde{\Psi} \,d\tilde{\mathrm{P}}_{n_\ell}
    \;\;\longrightarrow\;\;
    \int \tilde{\Psi} \,d\tilde{\mathrm{P}}_0^\prime
    \qquad(\ell\rightarrow\infty)
  \end{eqnarray} 
  for Pettis and Bochner integrals (since
  $H$ is assumed to be a separable Banach space,
  Pettis integrals and Bochner integrals coincide;
  see e.g.\ \citep[p.\ 150]{dudley1989}). 
  
  Let $H^\ast$ be the dual space of $H$. Note that 
  $F\circ\Psi$ is continuous and bounded on $\mathcal{Z}$
  for every $F\in H^\ast$. Hence,
  it follows from weak convergence of
  $(\mathrm{P}_{n_\ell})_{\ell\in\mathds{N}}$ 
  to $\mathrm{P}_0$ and a property of the Bochner integral
  \citep[Theorem 3.10.16]{denkowski2003} that
  $$\lim_{\ell\rightarrow\infty}
    F\bigg(\int \Psi\,d\mathrm{P}_{n_\ell}\bigg)
    =
    \lim_{\ell\rightarrow\infty}\int F\circ\Psi\,d\mathrm{P}_{n_\ell}
    =\int F\circ\Psi\,d\mathrm{P}_{0}
    =F\bigg(\int \Psi\,d\mathrm{P}_{0}\bigg) .
  $$
  Accordingly, vague convergence of 
  $(\tilde{\mathrm{P}}_{n_\ell})_{\ell\in\mathds{N}}$
  to 
  $\tilde{\mathrm{P}}_0^\prime$ implies
  $\lim_{\ell\rightarrow\infty}
    F\big(\int \tilde{\Psi}\,
          d\tilde{\mathrm{P}}_{n_\ell}
     \big)
   =F\big(\int \tilde{\Psi}\,
          d\tilde{\mathrm{P}}_0^\prime
     \big)
  $.
  Hence,
  \begin{eqnarray}\label{theorem-continuity-bochner-1003}
     \lim_{\ell\rightarrow\infty}
     F\bigg(\int \Psi\,d\mathrm{P}_{n_\ell}-
            \int \tilde{\Psi}\,d\tilde{\mathrm{P}}_{n_\ell}
      \bigg)
     \;=\;
     F\bigg(\int \Psi\,d\mathrm{P}_{0}-
            \int \tilde{\Psi}\,d\tilde{\mathrm{P}}_0^\prime
      \bigg)\;.
  \end{eqnarray}
  For every $\ell\in\mathds{N}$,
  \begin{eqnarray}\label{theorem-continuity-bochner-1005}
    \bigg\|
      \int\!\! \Psi\,d\mathrm{P}_{n_\ell}-\!\!
      \int\!\! \tilde{\Psi}\,d\tilde{\mathrm{P}}_{n_\ell}
    \bigg\|_H
    =
    \bigg\|
         \int_{\mathcal{Z}\setminus K}\! \Psi\,d\mathrm{P}_{n_\ell}
       \bigg\|_H  
       \leq\int_{\mathcal{Z}\setminus K} \|\Psi\|_H
               \,d\mathrm{P}_{n_\ell}
       \stackrel{(\ref{theorem-continuity-bochner-1})}{\leq}
           \frac{\varepsilon}{4}\,.\;
  \end{eqnarray}
  For every $\ell\in\mathds{N}$ and
  every $F\in H^\ast$ such that $\|F\|_{H^\ast}\leq 1$, 
  (\ref{theorem-continuity-bochner-1005}) 
  implies 
  $\big|F\big(\int \Psi\,d\mathrm{P}_{n_\ell}-
              \int \tilde{\Psi}\,d\tilde{\mathrm{P}}_{n_\ell}
         \big)
   \big|
   \,\leq\,\frac{\varepsilon}{4}
  $ and, because of (\ref{theorem-continuity-bochner-1003}), also
  $\big|F\big(\int \Psi\,d\mathrm{P}_{n_\ell}-
              \int \tilde{\Psi}\,d\tilde{\mathrm{P}}_{n_\ell}
         \big)
   \big|
   \,\leq\,\frac{\varepsilon}{4}
  $\,.    
  Hence, it follows from \cite[Corollary II.3.15]{dunford1958}
  that
  \begin{eqnarray}\label{theorem-continuity-bochner-1004}
    \bigg\|\int \Psi\,d\mathrm{P}_{0}-
            \int \tilde{\Psi}\,d\tilde{\mathrm{P}}_0^\prime
    \bigg\|_H
    \;\leq\;\frac{\varepsilon}{4}\;.
  \end{eqnarray}
  By using the triangle inequality, we obtain
  \begin{eqnarray*}
    \lefteqn{
      \bigg\|\int \Psi \,d\mathrm{P}_{n_\ell}-\int \Psi \,d\mathrm{P}_0
      \bigg\|_{H}
    }\\
    &\leq&\!\!\!
      \bigg\|\!\int\!\! \Psi \,d\mathrm{P}_{n_\ell}\!-\!\!
             \int\!\! \tilde{\Psi}\,d\tilde{\mathrm{P}}_{n_\ell}
      \bigg\|_{H}\!\!+
      \bigg\|\int\!\! \tilde{\Psi}\,d\tilde{\mathrm{P}}_{n_\ell}\!-\!\!
             \int\!\! \tilde{\Psi}\,d\tilde{\mathrm{P}}_0^\prime
      \bigg\|_{H}\!\!+
      \bigg\|\int\!\! \tilde{\Psi}\,d\tilde{\mathrm{P}}_0^\prime\!-\!\!
             \int\!\! \Psi \,d\mathrm{P}_0
      \bigg\|_{H}\!,
  \end{eqnarray*}
  so that (\ref{theorem-continuity-bochner-1002}),
  (\ref{theorem-continuity-bochner-1005})
  and (\ref{theorem-continuity-bochner-1004}) imply
  $\limsup_{\ell\rightarrow\infty}
       \left\|\int \Psi \,d\mathrm{P}_{n_\ell}
              -\int \Psi \,d\mathrm{P}_0\right\|_{H}
    \,\leq\,\frac{\varepsilon}{2}.  
  $
  This is a contradiction to (\ref{theorem-continuity-bochner-1001}). 
\hfill$\Box$

\medskip

\textbf{Proof of Proposition \ref{prop-counterexample}:}
  Without loss of generality, we may assume that
  \begin{eqnarray}\label{prop-counterexample-400-1}
    \tilde{f}(x_{0})=0\qquad\text{and}\qquad
      \tilde{f}(x_{1})=1\;.
  \end{eqnarray}
  (Otherwise, we can divide $\tilde{f}$ by $\tilde{f}(x_{1})$\,.)
  Since the function
  $\mathds{R}\rightarrow[0,\infty),\;\;t\mapsto L(x_{1},1,t)$
  is convex, it is also continuous.
  Therefore, (\ref{prop-counterexample-b}) implies the existence
  of an $\gamma\in(0,1)$ such that 
  \begin{eqnarray}\label{prop-counterexample-400-2}
    L(x_{1},1,\gamma)\;>\;0\;.
  \end{eqnarray}
  Note that convexity of the loss function,
  $L(x_{1},1,1)=0$ and 
  $L(x_{1},1,\gamma)>0$ imply
  \begin{eqnarray}\label{prop-counterexample-1}
    0\;=\;L(x_{1},1,1)\;\leq\;L(x_{1},1,t)\;<\;
    L(x_{1},1,\gamma)\;\leq\;
    L(x_{1},1,s)
  \end{eqnarray}
  for $0\leq s\leq \gamma < t\leq 1$\,. 
  Define $\mathrm{P}_{0}:=\delta_{(x_{0},0)}$\,. 
  Since $f_{L,\delta_{(x_{0},0)},\lambda_{n}}\,=\,0$\,, it follows that
  \begin{eqnarray}\label{prop-counterexample-2}
    \mathrm{P}_{0}^{n}
    \Big(
      \big\{D_{n}\in(\mathcal{X}\times\mathcal{Y})^{n}\;
      \big|\;\;f_{L,D_{n},\lambda_{n}}\,=\,0
      \big\}
    \Big)\;\;=\;\;1\;.
  \end{eqnarray}
  Next, fix any $\dddelta\in(0,1)$ and define the mixture distribution
  $$\mathrm{P}_{\dddelta}\;:=\;
    (1-\dddelta)\mathrm{P}_{0}+\dddelta\delta_{(x_{1},1)}
    \;=\;(1-\dddelta)\delta_{(x_{0},0)}+\dddelta\delta_{(x_{1},1)}\;.
  $$
  For every $n\in\mathds{N}$\,, let
  $\mathcal{Z}_{n}^{\prime}$ be the subset of 
  $(\mathcal{X}\times\mathcal{Y})^{n}$ which consists of all
  those elements 
  $D_{n}=\big(D_{n}^{(1)},\dots,D_{n}^{(n)}\big)
   \in(\mathcal{X}\times\mathcal{Y})^{n}
  $
  where
  $$D_{n}^{(i)}\,\in\,\big\{(x_{0},0),(x_{1},1)\big\}
    \qquad\forall\,i\in\{1,\dots,n\}\;.
  $$
  In addition, let $\mathcal{Z}_{n}^{\prime\prime}$ be the subset of 
  $(\mathcal{X}\times\mathcal{Y})^{n}$ which consists of all
  those elements 
  $D_{n}=\big(D_{n}^{(1)},\dots,D_{n}^{(n)}\big)
   \in(\mathcal{X}\times\mathcal{Y})^{n}
  $
  where 
  \begin{eqnarray}\label{prop-counterexample-102}
      \sharp\,\Big(\!\big\{i\in\{1,\dots,n\}\;
                 \big|\;\;D_{n}^{(i)}=(x_{1},1)
                 \big\}\!
            \Big)\;\geq\;\frac{\varepsilon}{2}\;.
  \end{eqnarray}
  Define
  $\mathcal{Z}_{n}:=
   \mathcal{Z}_{n}^{\prime}\cap \mathcal{Z}_{n}^{\prime\prime}
  $\,.
  Then, we have 
  $\mathrm{P}_{\dddelta}^{n}(\mathcal{Z}_{n}^{\prime})=1$ and,
  according to the law of large numbers
  (\citet[Theorem 8.3.5]{dudley1989}),
  $\lim_{n\rightarrow\infty}
   \mathrm{P}^{n}_{\dddelta}(\mathcal{Z}_{n}^{\prime\prime})=1
  $\,.
  Hence, there is an $n_{\dddelta,1}\in\mathds{N}$ such that
  \begin{eqnarray}\label{prop-counterexample-3}
    \mathrm{P}^{n}_{\dddelta}(\mathcal{Z}_{n})
    \;\geq\;\frac{1}{2}
    \qquad\forall\,n\geq n_{\dddelta,1}\;.
  \end{eqnarray}
  Due to $\lim_{n\rightarrow\infty}\lambda_{n}=0$ and
  (\ref{prop-counterexample-400-2}), there is an 
  $n_{\dddelta,2}\in\mathds{N}$ such that
  \begin{eqnarray}\label{prop-counterexample-4}
    \lambda_{n}\|\tilde{f}\|_{H}^{2}
    \;<\;\frac{\dddelta}{2}L(x_{1},1,\gamma)
    \qquad\forall\,n\geq n_{\dddelta,2}\;.
  \end{eqnarray}
  In the following, we show
  \begin{eqnarray}\label{prop-counterexample-5}
    f_{L,D_{n},\lambda_{n}}(x_{1})\;>\;\gamma
    \qquad\forall\,D_{n}\in\mathcal{Z}_{n}\,,
    \quad\forall\,n\geq n_{\dddelta,2}\;.
  \end{eqnarray}
  To this end, fix any $D_{n}\in\mathcal{Z}_{n}$\,. 
  In order to prove (\ref{prop-counterexample-5}), it is enough
  to show the following assertion for every $n\geq n_{\dddelta,2}$\,:
  \begin{eqnarray}\label{prop-counterexample-400-3}
    \;\;f\in H\,,\;\;f(x_{1})\leq\gamma
    \quad\;\Rightarrow\;\quad
    \mathcal{R}_{L,D_{n},\lambda_{n}}(\tilde{f})\;\leq\;
    \mathcal{R}_{L,D_{n},\lambda_{n}}(f)\;.
  \end{eqnarray}
  The definition of $\mathcal{Z}_{n}$ and
  (\ref{prop-counterexample-400-1}) imply 
  $$\mathcal{R}_{L,D_{n},\lambda_{n}}(\tilde{f})\;=\;
    \mathcal{R}_{L,D_{n}}(\tilde{f})\,+\,\lambda_{n}\|\tilde{f}\|_{H}^{2}
    \;=\;\lambda_{n}\|\tilde{f}\|_{H}^{2}\;\;.
  $$
  For every $f\in H$ such that $f(x_{1})\leq\gamma$, 
  the definition of $\mathcal{Z}_{n}$ implies
  $$\mathcal{R}_{L,D_{n},\lambda_{n}}(f)
      \;\geq\;\mathcal{R}_{L,D_{n}}(f)
      \;\stackrel{(\ref{prop-counterexample-102})}{\geq}\;
         \frac{\dddelta}{2}L\big(x_{1},1,f(x_{1})\big)
      \;\stackrel{(\ref{prop-counterexample-1})}{\geq}\;
         \frac{\dddelta}{2}L\big(x_{1},1,f(x_{1})\big)\;.
  $$
  Hence, (\ref{prop-counterexample-400-3}) follows from
  (\ref{prop-counterexample-4}) and, therefore,
  we have proven (\ref{prop-counterexample-5}).

  Define $n_{\varepsilon}=\max\{n_{\varepsilon,1},n_{\varepsilon,2}\}$\,.
  By assumption, $k$ is a bounded, non-zero kernel. 
  According to \citet[Lemma 4.23]{steinwart2008}, 
  this implies
  $$\|f_{L,D_{n},\lambda_{n}}\|_{H}\;\geq\;
    \frac{\,\|f_{L,D_{n},\lambda_{n}}\|_{\infty}}{\|k\|_{\infty}}
    \;\stackrel{(\ref{prop-counterexample-5})}{\geq}\;
    \frac{\gamma}{\|k\|_{\infty}}
    \qquad\forall\,D_{n}\in\mathcal{Z}_{n}\,,
    \quad\forall\,n\geq n_{\dddelta}
  $$
  and, therefore,
  \begin{eqnarray}\label{prop-counterexample-6}
    \qquad\|f_{L,D_{n},\lambda_{n}}\|_{H}\;\geq\;
    \min\left\{\frac{\gamma}{\|k\|_{\infty}}\,,\,1\right\}\;=:\;c
    \quad\;\;\forall\,D_{n}\in\mathcal{Z}_{n}\,,
    \;\;\forall\,n\geq n_{\dddelta}\,.
  \end{eqnarray}
  Define $F:=\{f\in H\,|\,\,\|f\|_{H}\geq c\}\,$ and
  \begin{eqnarray}\label{prop-counterexample-7}
    \qquad F^{\frac{c}{2}}:=
    \big\{f\in H\,
    \big|\,\,\inf_{f^{\prime}\in H}\|f-f^{\prime}\|_{H}
             \leq{\textstyle\frac{c}{2}}
    \big\}
    \;\subset\;
    \big\{f\in H\,\big|\;\|f\|_{H}>0\big\}\;.
  \end{eqnarray}
  Hence, for every $n\geq n_{\dddelta}$\,, we obtain
  \begin{eqnarray*}
    \lefteqn{\big[\SVMn_{n}(\mathrm{P}_{\dddelta}^{n})\big](F)
      \;=\;\mathrm{P}_{\dddelta}^{n}
             \Big(\big\{D_{n}\,
                  \big|\,\|f_{L,D_{n},\lambda_{n}}\|_{H}\geq c
                  \big\}
             \Big) 
      \;\stackrel{(\ref{prop-counterexample-6})}{\geq}\;
        \mathrm{P}_{\dddelta}^{n}(\mathcal{Z}_{n}) } \\
    &&\stackrel{(\ref{prop-counterexample-3})}{\geq}\;
        \frac{1}{2}
      \;\stackrel{(\ref{prop-counterexample-6})}{\geq}\;\frac{\,c}{\,2}
      \;\stackrel{(\ref{prop-counterexample-2})}{=}\;
      \mathrm{P}_{0}^{n}
             \Big(\big\{D_{n}\,
                  \big|\,\|f_{L,D_{n},\lambda_{n}}\|_{H}> 0
                  \big\}
             \Big)\,+\,\frac{\,c}{\,2} \qquad \\
    &&=\;\big[\SVMn_{n}(\mathrm{P}_{0}^{n})\big]
          \Big(\big\{f\in H\;\big|\;\|f\|_{H}>0\big\}\Big)
          \,+\,\frac{\,c}{\,2} \\
    &&\stackrel{(\ref{prop-counterexample-7})}{\geq}\;
        \big[\SVMn_{n}(\mathrm{P}_{0}^{n})\big]
          \Big(F^{\frac{c}{2}}\Big)
          \,+\,\frac{\,c}{\,2}\;.
  \end{eqnarray*}
  According to the definition of the Prokhorov distance
  (see Subsection \ref{sec-weak-convergence-bochner}),
  it follows that
  \begin{eqnarray}\label{prop-counterexample-8} 
    \sup_{n\in\mathds{N}}\,
    d_{\textrm{Pro}}
     \Big(\SVMn_{n}(\mathrm{P}_{0}^{n}),\SVMn_{n}(\mathrm{P}_{\dddelta}^{n})
     \Big)
    \;\geq\;\frac{\,c}{\,2}
  \end{eqnarray}
  In addition, we have 
  $d_{\textrm{Pro}}\big(\mathrm{P}_{0},\mathrm{P}_{\dddelta}\big)
    \leq\dddelta
  $
  because $\mathrm{P}_{\dddelta}$ is an $\varepsilon$-mixture of
  $\mathrm{P}_{0}$\,.
  Since $c>0$ does not depend on $\dddelta\in(0,1)$
  and $\dddelta$ may be arbitrarily small, this proves
  that $(\SVMn_{n})_{n\in\mathds{N}}$ is not
  qualitatively robust in $\mathrm{P}_{0}$\,.
\hfill$\Box$

\bibliographystyle{abbrvnat}
\bibliography{literatur}

\end{document}